\documentclass[journal]{IEEEtran}
\usepackage{amsmath,graphicx}
\usepackage{subfigure}
\usepackage{amsfonts}
\usepackage{xcolor}
\usepackage{hyperref}
\usepackage{xr}

\begin{document}

\title{Deep Generalization of Structured Low-Rank Algorithms (Deep-SLR)  }

\author{\IEEEauthorblockN{Aniket Pramanik,
Hemant Aggarwal,
Mathews Jacob 
}\\
\IEEEauthorblockA{The University of Iowa, USA}

\thanks{Aniket Pramanik, Hemant Aggarwal and Mathews Jacob are from the Department of Electrical and Computer Engineering at the University of Iowa, Iowa City, IA, 52242, USA (e-mail: aniket-pramanik@uiowa.edu; hemantkumar-aggarwal@uiowa.edu; mathews-jacob@uiowa.edu). This work is supported by grant NIH 1R01EB019961-01A1.}}

\IEEEtitleabstractindextext{%
\begin{abstract}
	Structured low-rank (SLR) algorithms, which exploit annihilation relations between the Fourier samples of a signal resulting from different properties, is a powerful image reconstruction framework in several applications. This scheme relies on low-rank matrix completion to estimate the annihilation relations from the measurements. The main challenge with this strategy is the high computational complexity of matrix completion. We introduce a deep learning (DL) approach to significantly reduce the computational complexity. Specifically, we use a convolutional neural network (CNN)-based filterbank that is trained to estimate the annihilation relations from imperfect (under-sampled and noisy) k-space measurements of Magnetic Resonance Imaging (MRI). 
The main reason for the computational efficiency is the pre-learning of the parameters of the non-linear CNN from exemplar data, compared to SLR schemes that learn the linear filterbank parameters from the dataset itself.
Experimental comparisons show that the proposed scheme can enable calibration-less parallel MRI; it can offer performance similar to SLR schemes while reducing the runtime by around three orders of magnitude. Unlike pre-calibrated and self-calibrated approaches, the proposed uncalibrated approach is insensitive to motion errors and affords higher acceleration. The proposed scheme also incorporates image domain priors that are complementary, thus significantly improving the performance over that of SLR schemes.
\end{abstract}

}

\maketitle

\IEEEdisplaynontitleabstractindextext

%
\IEEEpeerreviewmaketitle

\section{Introduction}
Parallel MRI was introduced to speed up the traditionally slow MR acquisitions.  Specifically, the redundancy between the k-space samples is capitalized to highly undersample k-space data, thereby reducing the scan time. The recovery of images from highly under-sampled multi-channel Fourier measurements is a classical problem in MRI \cite{lustig2007sparse}. Pre-calibrated approaches such as SENSE \cite{pruessmann1999sense} rely on coil sensitivities that are estimated using additional calibration scans.  Several image priors \cite{doneva2020mathematical}, including sparsity, have been used to regularize pre-calibrated image recovery, resulting in improved image recovery at high acceleration factors. Researchers have recently introduced model-based deep-learning (DL) algorithms that use a forward model (capturing the imaging physics) combined with a deep-learned prior \cite{sun2016deep,schlemper2017deep,hammernik2018learning,aggarwal2018modl}. An improvement in image quality as a result of multiple iterations of optimization blocks, sharing weights across the network, and end-to-end training, along with the ability to use multiple learned regularization priors, has been demonstrated in \cite{aggarwal2018modl}. Since these methods use pre-estimated coil sensitivity maps within the forward model, they suffer from errors in the sensitivity maps resulting from motion or high accelerations. Self-calibrated approaches such as GRAPPA \cite{griswold2002generalized}, SPIRiT \cite{lustig2010spirit} and ESPIRiT \cite{uecker2014espirit} estimate coil sensitivities from a fully sampled calibration region in the center of k-space. However, the need for a fully sampled region restricts the achievable acceleration in these settings.

Structured low-rank (SLR) matrix completion approaches \cite{jacob2020structured,haldar2019linear, haldar2014low} were introduced to overcome the challenges with the above calibration based schemes and have been very effective in uncalibrated parallel MRI \cite{uecker2014espirit} and multi-shot acquisitions \cite{mani2017multi}. In the context of parallel MRI, these methods exploit the annihilation relations between multi-channel Fourier data rather than relying on explicit coil sensitivity estimates. Similar SLR approaches have been used to exploit a variety of other signal properties, including support constraints\cite{haldar2019linear}, continuous domain sparsity \cite{ongie2015super, lee2016acceleration},  phase \cite{haldar2019linear}, and the exponential structure of an MRI time series \cite{jacob2020structured}. Iterative re-weighted least-squares (IRLS) SLR algorithms make use of the convolutional structure of the matrices \cite{ongie2017fast} to accelerate the algorithms. IRLS methods alternate between estimating an annihilation (null space) filterbank and updating the Fourier coefficients of the signal from the available measurements. Specifically, the missing Fourier coefficients are chosen so they match the measurements while being annihilated by the filterbank; the projection energy of the signal to the signal subspace measured by a residual convolution-deconvolution filterbank, is maximized. While this algorithm is considerably faster than earlier approaches, the iterative estimation of the annihilation filterbank from the under-sampled data is still computationally expensive. The approaches that use calibration information \cite{uecker2014espirit,ongie2015super,haldar2019linear} estimate the null space filters from a fully sampled calibration region, resulting in reduced complexity. Since the  annihilation filterbank need not be derived from the under-sampled data in an iterative fashion, this approach offers faster reconstructions. However, the challenge with these methods is the need for a calibration region, which restricts the achievable acceleration.

In this paper, we introduce a general DL strategy to reduce the runtime of  the SLR algorithms, which is valid for all the signal priors discussed above \cite{ongie2015super, lee2016acceleration,haldar2019linear,jacob2020structured}. Unlike the SLR approach that estimates a specific linear annihilation network for each dataset from the undersampled measured data, we propose to learn a single non-linear CNN from several training datasets. Specifically, the residual convolution-deconvolution linear filterbank in the IRLS-SLR algorithm is replaced with a residual multi-channel CNN. We hypothesize that the pre-learned non-linear CNN behaves as a different linear annihilation filterbank for each specific dataset, annihilating the multi-channel data. The residual CNN behaves as a projection for each dataset, facilitating the \emph{denoising} of the dataset from alias artifacts and noise at each iteration. Similar to MoDL \cite{aggarwal2018modl}, the proposed model unrolls the resulting algorithm and learns the parameters of the non-linear filterbank in an end-to-end fashion. The proposed work combines this approach with an image domain prior similar to MoDL, which is complementary to the Fourier domain multi-channel relations. This hybrid approach offers improved performance over SLR, while offering around three orders of magnitude reduction in computational complexity. We focus on two representative applications---sparse single-channel recovery and parallel MRI---which use two distinct lifting structures in the SLR approach \cite{jacob2020structured}. Specifically, we show how different lifting structures can be accommodated in the proposed scheme by modifying the data organization of the input and output of the CNN module. This enables the extension of the proposed framework to a range of SLR applications \cite{jacob2020structured,haldar2019linear,shin2014calibrationless} that use one or a combination of the above lifting structures. 


This main focus of this work is to introduce a general DL framework for uncalibrated parallel MRI and multishot MRI. This work is related to \cite{aggarwal2019modl} and \cite{hu2019reconstruction}, which are partially calibrated strategies. Specifically, the MoDL-MUSSELS \cite{aggarwal2019modl} framework explicitly accounted for the pre-estimated coil sensitivities within the data consistency block, while it performed a calibration-free correction of phase errors between shots. A challenge with these partially calibrated approaches is the potential mismatch between coil sensitivities and the diffusion-weighted acquisition due to motion between the coil sensitivity calibration scan and the diffusion scan. By contrast, the annihilation relation between coils is learned by the non-linear k-space CNN from exemplar data in this work; the application of this framework to the diffusion setting yields a completely uncalibrated algorithm, which jointly accounts for coil sensitivities as well as phase errors between shots. As demonstrated in our experiments in the supplementary material, this approach eliminates errors resulting from a motion-induced mismatch between the calibration scan and the diffusion-weighted one. In addition, the main focus of this work is to show that the proposed approach works well for a range of SLR priors, of which only one was considered in the earlier work \cite{aggarwal2019modl}. The conference version of this work is presented in the literature \cite{pramanik2019off, pramanik2020calibrationless}. DL methods in k-space were the focus of recent work \cite{han2019k,eo2018kiki,akccakaya2019scan}. The RAKI framework \cite{akccakaya2019scan} is a calibrated scheme, unlike our calibration-free approach. A direct inversion (model-free) approach was pursued in one study \cite{han2019k}; it differs from the proposed model-based framework, which also combines image domain priors. The KIKI net approach \cite{eo2018kiki} was introduced for a single-channel setting, unlike our uncalibrated multi-channel scheme.  The proposed reconstructions are compared against model-free image domain DL \cite{ye2018deep}, k-space DL \cite{han2019k}, image domain MoDL \cite{aggarwal2018modl}, and traditional SLR methods. These comparisons reveal the improved performance offered by the Deep-SLR framework. 
\section{Background}
We now briefly describe the background to make the paper self-contained and easily accessible. 
\subsection{Forward Model}
We model the acquisition of image ${\boldsymbol \gamma}(\mathbf r)$ as: 
\begin{equation}
\label{forward_model}
  b_i = \mathcal S (\mathcal{F}(\underbrace{ s_i~{\boldsymbol \gamma}}_{\boldsymbol \gamma_i})) + {\eta_i},\hspace{2pt} i=1\ldots M,
\end{equation}
where $ s_i; i=1,..,M$ is the coil sensitivity of the $i^{\rm th}$ coil, while $b_i$ is the noisy under-sampled Fourier measurement, and $\boldsymbol \gamma_i$ is the image, corresponding to the $i^{\rm th}$ coil. $\eta_i$ is the noise term.
Here $\mathcal F$ is the Fourier transform that maps $\boldsymbol \gamma_i$ onto its k-space samples and $\mathcal S$ is the under-sampling operator. We compactly denote the above operation as 
\begin{equation}
\mathbf B = \mathcal{A}({\boldsymbol \Gamma}) + \mathbf P
\end{equation} 
where $\widehat{\boldsymbol{\Gamma}} = \left [\begin{array}{ccc}
{\widehat{{\boldsymbol\gamma}}_1} & .. & \widehat{{\boldsymbol\gamma}}_M 
\end{array}\right]$ is the matrix representing multi-channel data in Fourier space, $ \mathbf B = \left [\begin{array}{ccc}
\mathbf{b}_1 & .. & \mathbf{b}_M 
\end{array}\right]$ is the corresponding noisy under-sampled multi-channel Fourier measurement, and $\mathbf P=\left [\begin{array}{ccc}
\boldsymbol{\eta}_1 & .. & \boldsymbol{\eta}_M 
\end{array}\right]
$ is the multi-channel noise. Note that we denote the image of the $i^{\rm th}$ channel by $\boldsymbol \gamma_i$, while $\boldsymbol{\Gamma}$ denotes the concatenation of the channel data in spatial domain. The Fourier domain representation of $i^{\rm th}$ channel image is $\widehat{\boldsymbol \gamma_i}$, while $\widehat{\boldsymbol \Gamma}$ is the concatenation of the channel images in Fourier domain. 
\subsection{Structured Low-Rank Algorithms}
\label{slrbackground}
SLR methods rely on different liftings of the Fourier coefficients, designed to exploit specific properties of the signal. We now discuss two representative SLR applications, which illustrate the different types of lifting used in the SLR setting.
\subsubsection{Continuous domain sparsity} 
\label{girafsec}
A continuous domain piecewise constant image $\boldsymbol \gamma$ with edges specified by the zero sets of a bandlimited function $\mu$ satisfies an image domain annihilation relation, $\nabla \boldsymbol \gamma(\mathbf r) \cdot \mu(\mathbf r)=0, \forall \mathbf r $, where $\mathbf r$ represents spatial coordinates. Here, $\nabla \boldsymbol \gamma$ denotes the gradient of $\boldsymbol \gamma$. This relation translates to the following Fourier domain annihilation relations $\widehat{\nabla \boldsymbol \gamma}[\mathbf k] \ast n[\mathbf k]=0, \forall \mathbf k $, where $\mathbf k $ denotes k-space coordinates (Fourier space). Here $\widehat{\nabla \boldsymbol \gamma}[\mathbf k]$ represents the Fourier coefficients of the gradient of $\boldsymbol \gamma$ and $n[\mathbf k]$ is the Fourier transform of $\mu(\mathbf r)$. We denote the mapping from the Fourier coefficients ${\widehat{\boldsymbol \gamma}}$ to $\widehat{\nabla \boldsymbol \gamma}$ by $\mathcal G$:
\begin{equation}
\label{nabla_def}
\mathcal G({\widehat{\boldsymbol \gamma}}) = 
\widehat{\nabla \boldsymbol \gamma}[\mathbf k]=\left[\begin{array}{c} j2\pi {k_x}~ \widehat{ \boldsymbol \gamma}[\mathbf k] \\ j2 \pi {k_y}~ \widehat{\boldsymbol \gamma}[\mathbf k] \end{array} \right ]= \left[\begin{array}{c}  \widehat{\boldsymbol{\gamma}_x} \\  \widehat{\boldsymbol{ \gamma}_y} \end{array} \right ]. 
\end{equation}
Note that $\mathcal G$ essentially creates two copies of $\widehat{\boldsymbol \gamma}$, each with a different Fourier weighting.

The convolution relation $\widehat{\nabla \boldsymbol \gamma}[\mathbf k] \ast n[\mathbf k]=0$ can be represented as Hankel matrix multiplication $\mathcal{H}(\widehat{\nabla \boldsymbol \gamma})~\mathbf n=0$. The number of such null space filters, denoted by $V$, is often large (see \cite{ongie2017convex}) 
\begin{equation}\label{gradient}
\underbrace{\begin{bmatrix}
\mathcal H(\widehat{\boldsymbol \gamma_x})\\\mathcal H(\widehat{\boldsymbol \gamma_y})
	\end{bmatrix}}_{\mathcal T\left(\mathcal G(\widehat{\boldsymbol \gamma})\right) }\underbrace{\left[\begin{array}{c|c|c|c}
\mathbf n_1& \mathbf n_2 &\ldots & \mathbf n_V
\end{array}\right]}_{\mathbf N} = 0.
\end{equation}
resulting in a low-rank matrix $\mathcal T(\mathcal G(\widehat{\boldsymbol \gamma}))$. 

Note that the Hankel matrices are vertically stacked to obtain $\mathcal T\left(\mathcal G(\widehat{\boldsymbol \gamma})\right)$, which is a common approach in SLR \cite{uecker2014espirit}.
\subsubsection{Parallel MRI acquisition scheme}
\label{pslrsec}
Image and Fourier domain multi-channel annihilation relations were shown in two studies \cite{uecker2014espirit,jacob2020structured}. Specifically, each pair of multi-channel images in \eqref{forward_model} satisfy a Fourier domain annihilation relation
$\widehat{\boldsymbol \gamma_{i}}[\mathbf{k}]\ast\widehat{\boldsymbol s_{j}}[\mathbf{k}]-\widehat{\boldsymbol \gamma_{j}}[\mathbf{k}]\ast\widehat{\boldsymbol s_{i}}[\mathbf{k}]=0, \forall \mathbf{k}$, where $\widehat{\boldsymbol \gamma_{i}}[\mathbf{k}]$ and $\widehat{\boldsymbol s_{i}}[\mathbf{k}]$ are the Fourier coefficients of $\boldsymbol \gamma_{i}(\mathbf{k})$ and $\boldsymbol s_{i}(\mathbf{k})$, respectively. Such annihilation relations exist for every pair of coil images and can be compactly written as
\begin{eqnarray}
\label{compact_rel}
\underbrace{\left[\begin{array}{cccc}\mathcal{H} (\widehat{\boldsymbol \gamma_{1}})
& \mathcal{H} (\widehat{\boldsymbol \gamma_{2}}) & \ldots & \mathcal{H} (\widehat{\boldsymbol \gamma_{M}})\end{array}\right]}_{\mathcal{T}({\widehat{\boldsymbol \Gamma}})}\cdot \mathbf{N} = 0.
\end{eqnarray}
The columns of $\mathbf N$ correspond to the vertical stacking of the filters $\widehat{\boldsymbol s_{i}}$. The large null space $\mathbf{N}$ implies it is low rank. Note that the Hankel matrices are horizontally stacked to obtain $\mathcal T\left(\widehat{\boldsymbol{\Gamma}}\right)$. Here $\mathcal G=\mathcal I$, which is the identity mapping. This is another popular class of lifting used in SLR \cite{jacob2020structured,haldar2019linear,lee2016acceleration}. 
\subsection{Calibration-free SLR Methods}
In general, SLR schemes aim to recover an image or a series of images $\boldsymbol \Gamma$ from its measurements $\mathcal A(\boldsymbol \Gamma)$ by solving the optimization problem:
\begin{equation}
\label{slr}
\min_{{\boldsymbol \Gamma}}\hspace{2pt} \mbox{rank}\hspace{1pt} \big[\mathcal{T}(\mathcal{G}(\widehat{\boldsymbol \Gamma}))\big] \hspace{2pt} \mbox{such that} \hspace{2pt} \mathbf B=\mathcal A\left( {\boldsymbol \Gamma}\right)+\mathbf P.
\end{equation}
Here, $\mathcal T(.)$ is a lifting operator that lifts the weighted signal $\mathcal{G}(\widehat{\boldsymbol \Gamma})$ into a higher dimensional structured matrix. As discussed earlier, the generic weighting matrix $\mathcal G$ depends on the specific annihilation relation. The recovery of ${\boldsymbol \Gamma}$ is often posed as an unconstrained nuclear norm minimization problem
\begin{equation} 
\label{slr1}
\arg \min_{{\boldsymbol \Gamma}} \|\mathcal{A}({\boldsymbol \Gamma})-\mathbf B\|_2^2 + \lambda \|\mathcal{T}(\mathcal G (\widehat{\boldsymbol \Gamma}))\|_\ast
\end{equation}
where $\lambda$ is a regularizer to tune the nuclear norm loss term. 
\subsection{Iterative Re-weighted Least-Squares (IRLS) Algorithm}
\label{irls}
The IRLS scheme majorizes the nuclear norm with a weighted Frobenius norm as $\|\mathcal T(\mathcal G (\widehat{\boldsymbol \Gamma}))\|_\ast \leq \|\mathcal T(\mathcal G (\widehat{\boldsymbol \Gamma}))\mathbf Q\|_F^2$ to yield a two-variable optimization problem
\begin{equation}
\label{constrained}
\arg \min_{{\boldsymbol \Gamma}, \mathbf Q} \|\mathcal{A}({\boldsymbol \Gamma})-\mathbf B\|_2^2 + \lambda \|\mathcal{T}(\mathcal G (\widehat{\boldsymbol \Gamma}))\mathbf Q\|_F^2,
\end{equation}
which alternates between the null space $\mathbf Q$ and image ${\boldsymbol \Gamma}$,
\begin{eqnarray}
\label{fup}
{\boldsymbol \Gamma}^{(n)}=\arg \min_{{\boldsymbol \Gamma}} \|\mathcal A(\mathbf{{\boldsymbol \Gamma}})-\mathbf B\|_2^2+ \lambda \|\mathcal T(\mathcal{G} (\widehat{\boldsymbol \Gamma}))\mathbf Q^{(n-1)}\|_F^2\\
\label{qup}
\mathbf Q^{(n)}= [\mathcal T(\mathcal G (\widehat{\boldsymbol \Gamma}^{(n)}))^H\mathcal T(\mathcal G(\widehat{\boldsymbol \Gamma}^{(n)}))+\epsilon^{(n)}\mathbf I]^{-1/4}
\end{eqnarray} respectively.   
The matrix $\mathbf Q$ can be viewed as a collection of column vectors spanning the null space of $\mathcal T(\mathcal G (\widehat{\boldsymbol \Gamma}))$. 
\subsection{Calibration-based SLR Methods}
\label{calibrated}
Several calibration-based MRI schemes (e.g., GRAPPA, SPIRiT \cite{griswold2002generalized,lustig2010spirit}) are related to the SLR schemes \cite{uecker2014espirit,jacob2020structured}. These approaches acquire a fully sampled calibration region in the Fourier domain, which corresponds to fully sampled rows of $\mathcal T(\mathcal G (\widehat{\boldsymbol{\Gamma}}))$ or, equivalently, the sub-matrix $\mathcal T_{R}(\mathcal G (\widehat{\boldsymbol{\Gamma}}))$. These schemes estimate the null space matrix $\mathbf Q$ (or, equivalently, the GRAPPA weights) by solving $\mathcal T_{R}(\mathcal G (\widehat{\boldsymbol{\Gamma}})) \mathbf Q =0$ subject to norm constraints on $\mathbf Q$; see the literature \cite{jacob2020structured} for details. 

Once the $\mathbf Q$ is pre-estimated from calibration data, the image is recovered from under-sampled Fourier coefficients by minimizing 
\begin{equation}
\label{lp}
\arg \min_{{\boldsymbol \Gamma}} \|\mathcal A(\mathbf{{\boldsymbol \Gamma}})-\mathbf B\|_2^2+ \lambda \|\mathcal{T}(\mathcal G (\widehat{\boldsymbol \Gamma}))\mathbf Q\|_F^2.
\end{equation}
The above optimization problem simplifies solving the system of equations $\mathcal A(\mathbf{{\boldsymbol \Gamma}})=\mathbf B; \mathcal G (\widehat{\boldsymbol \Gamma}))\mathbf Q=0$ for specific sampling patterns analytically \cite{griswold2002generalized,uecker2014espirit}. In other cases \cite{ongie2015super,haldar2019linear}, \eqref{lp} is solved iteratively. Both strategies are computationally efficient since $\mathbf Q$ is fully known. However, the need for a calibration region restricts the achievable acceleration.
\section{Deep Generalization of SLR Methods}
The main focus of this work is to introduce a DL solution to improve the computational efficiency of SLR algorithms. We note that calibrated SLR methods, which learn the linear null space projection operator from calibration data, require few iterations for convergence, thus offering fast image recovery. Calibration-free SLR methods by contrast are computationally expensive. Specifically, because the null space matrix $\mathbf Q$ is estimated from the data itself, the algorithm requires several iterations to converge. 

We propose to pre-learn a CNN-based null space projector from multiple exemplar datasets. The proposed non-linear CNN module learns to estimate the annihilation relations from the under-sampled data based on its training on exemplar data. We view this approach as learning a non-linear filterbank, which behaves like different linear filterbanks for different images. Specifically, the non-linear filterbank can be approximated as a linear filterbank, which projects the data to the null space, thus annihilating the signal but preserving the noise and alias artifacts; the residual block preserves the signal, while suppressing noise. 
\subsection{IRLS Algorithm with Variable Splitting}
To facilitate the reinterpretation of the reconstruction scheme as an iterative denoising strategy, we introduce an auxiliary variable $\widehat{\mathbf z}$ in \eqref{constrained} to obtain a three-variable constrained optimization problem,
\begin{equation}\label{constrainedeq}
\arg \min_{{\boldsymbol \Gamma}, \mathbf Q, \widehat{\mathbf{Z}}} \|\mathcal{A}({\boldsymbol \Gamma})-\mathbf B\|_2^2 + \lambda \|\mathcal{T}(\widehat{\mathbf{Z}})\mathbf Q\|_F^2 \hspace{3pt} \mbox{such that} \hspace{3pt} \widehat{\mathbf Z} = \mathcal G (\widehat{\boldsymbol \Gamma}).
\end{equation}
We impose the constraint by a penalty term as 
\begin{equation*}
\arg \min_{{\boldsymbol \Gamma}, \mathbf Q, \widehat{\mathbf{Z}}} \|\mathcal{A}({\boldsymbol \Gamma})-\mathbf B\|_2^2 + \lambda \|\mathcal{T}(\widehat{\mathbf{Z}})\mathbf Q\|_F^2+ \beta \|\mathcal G (\widehat{\boldsymbol \Gamma}) - \widehat{\mathbf{Z}}\|_2^2.
\end{equation*}
This formulation is equivalent to \eqref{constrainedeq} when $\beta \rightarrow \infty$.  We propose to solve the above problem using the alternating minimization scheme:
\begin{eqnarray}
\label{img_update}
\mathbf {{\boldsymbol \Gamma}}_{n+1} &=& \arg \min_{{\boldsymbol \Gamma}}  \|\mathcal A({\boldsymbol \Gamma}) - \mathbf B\|^2 + \beta \|\mathcal G (\widehat{\boldsymbol \Gamma}) -  \widehat{\mathbf Z}_{n}\|^2
\\\nonumber
\widehat{\mathbf Z}_{n+1} &=& \arg \min_{\widehat{\mathbf Z}} \beta \| \mathcal G (\widehat{\boldsymbol\Gamma}_{n+1}) - \widehat{\mathbf Z}\|^2+ \lambda \|\mathcal{T}(\widehat{\mathbf{Z}})\mathbf Q\|_F^2. \\\label{z_update}
\end{eqnarray} 
At each step, the $\mathbf Q$ matrix is updated as in \eqref{qup}.
\subsubsection{Image Update}The first step specified by \eqref{img_update} is a simple Tikhonov regularized optimization problem to recover the multi-channel images $\boldsymbol \gamma$ at the $(n+1)$-th iteration. When $\mathcal G=\mathcal I$, the prior reduces to $\|\widehat{\boldsymbol \Gamma}-  \widehat{\mathbf Z}_{n}\|^2$. In the general case, the solution to this optimization problem can be determined analytically as 
\begin{equation}
\widehat{\boldsymbol \Gamma}_{n}=(\mathcal{A}^H \mathcal A + \beta \mathcal G^H\mathcal G)^{-1}(\mathcal{A}^H \mathbf B + \beta \mathcal G^H(\widehat{\mathbf Z}_{n-1})),
\end{equation}
when $\mathcal A$ involves a sampling in the Fourier domain. Similar analytical solutions can also be used when $\mathcal G$ involves a Fourier domain weighting as in the literature \cite{gregSIAM2016}. 
\subsubsection{Projection }The sub-problem \eqref{z_update} is essentially a proximal operation. Specifically, the second term of  \eqref{z_update} is the energy in projecting $\mathcal T(\widehat{\mathbf Z})$ to the subspace $\mathbf Q$. If $\lambda \rightarrow \infty$, we obtain $\widehat{\mathbf Z}$ as the projection of $\widehat{\boldsymbol \Gamma}_{n+1}$ onto the signal subspace, orthogonal to $\mathbf Q$. 

\subsection{Filterbank Interpretation of the Denoising Subproblem}
We will now focus on the denoising sub-problem by showing its linear filterbank structure. We will capitalize on this structure to generalize the algorithm. 
We will focus on the vertical and horizontal stacking cases separately. 
\subsubsection{Vertical stacking considered in Section \ref{girafsec}}
Consider the term $\mathcal T(\widehat{\mathbf Z})\mathbf q_i$, where $\mathbf q_i$ is one of the columns of the matrix $\mathbf Q$. When the lifting operation is described by \eqref{gradient}, we have 
\begin{equation}
\mathcal T(\widehat{\mathbf Z})~\mathbf q_i = \begin{bmatrix}
\mathcal H(\widehat{\mathbf z}_1) \\\mathcal H(\widehat{\mathbf z}_2)\end{bmatrix}\mathbf q_i= \begin{bmatrix}
\mathbf p_1\\\mathbf p_2\end{bmatrix}
\end{equation} 
Because $\mathcal H(\widehat{\mathbf z})$ is a Hankel matrix, $\mathcal H(\widehat{\mathbf z})\mathbf Q$ corresponds to the linear convolution between $\widehat{\mathbf z}$ and $\mathbf Q$. Since convolution is commutative, we can rewrite the above expression as 

\begin{equation}
\mathcal T(\widehat{\mathbf Z})\mathbf q_i = \underbrace{\begin{bmatrix}
\widehat{\mathbf z}_1 \\\widehat{\mathbf z}_2\end{bmatrix}}_{\widehat{\mathbf Z}}\mathcal P(\mathbf q_i),
\end{equation} 
where, $\mathcal P(\mathbf q_i)$ is a block Hankel matrix constructed from the samples of $\mathbf q_i$. We thus have $\|\mathcal T(\widehat{\mathbf Z})\mathbf Q\|^2 = \| \widehat{\mathbf Z}\, \mathcal J(\mathbf Q)\|^2$, where $\mathcal J(\mathbf Q)$ is obtained by horizontally stacking the matrices $\mathcal P(\mathbf q_i)$. We note that $\widehat{\mathbf z}_1  J(\mathbf Q)$ corresponds to passing $\widehat{\mathbf z}_1$ through a single input multiple output (SIMO) filterbank, whose filters are specified by $\mathbf q_i$.

\subsubsection{Horizontal stacking considered in Section \ref{pslrsec}} Similar to the vertical stacking case, we consider
\begin{eqnarray}
\mathcal T(\widehat{\mathbf Z})~\mathbf q_i &=& \overbrace{\begin{bmatrix}
\mathcal H(\widehat{\mathbf z}_1) &..&\mathcal H(\widehat{\mathbf z}_N)\end{bmatrix}}^{\mathcal T(\widehat{\mathbf Z})} \overbrace{\begin{bmatrix}
\mathbf q_{i,1}\\\vdots\\\mathbf q_{i,N}\end{bmatrix}}^{\mathbf q_i}\\
&=&\underbrace{\begin{bmatrix}
\mathcal P(\mathbf q_{i,1}) &..&\mathcal P(\mathbf q_{i,N})\end{bmatrix}}_{\mathcal J(\mathbf Q)} \underbrace{\begin{bmatrix}
\widehat{\mathbf z}_{1}\\\vdots\\\widehat{\mathbf z}_{N}\end{bmatrix}}_{\widehat{\mathbf Z}}
\end{eqnarray} 
We thus have $\|\mathcal T(\widehat{\mathbf Z})\mathbf Q\|^2  = \|\mathcal J(\mathbf Q) \widehat{\mathbf Z}\|^2$, where 
\begin{equation}\label{key}
\mathcal J(\mathbf Q) = \begin{bmatrix}
\mathcal P(\mathbf q_{1,1}) &..&\mathcal P(\mathbf q_{1N})\\
\vdots&..&\vdots\\
\mathcal P(\mathbf q_{N,1}) &..&\mathcal P(\mathbf q_{N,N})
\end{bmatrix}
\end{equation}
We note that $\mathcal J(\mathbf Q)\widehat{\mathbf Z}$ corresponds to passing $\widehat{\mathbf Z}$ through a multiple input multiple output (MIMO) filterbank, whose filters are specified by $\mathbf q_i$. 
\subsection{Approximation of Denoising Sub-problem}
We thus rewrite \eqref{z_update} for both lifting approaches as,
\begin{equation}
\label{two_var_unconstrained}
\widehat{\mathbf Z}_{n+1} = 
\arg \min_{\widehat{\mathbf{Z}}} \beta \| \mathcal G (\widehat{\boldsymbol\Gamma}_{n+1}) - \widehat{\mathbf Z}\|^2 + \lambda \|\mathcal{J}(\mathbf Q_n)\widehat{\mathbf Z}\|_F^2.
\end{equation}  
which reduces to
\begin{equation}
\widehat{\mathbf Z}_{n+1} = \left[\mathbf I ~+~\frac{\lambda}{\beta}~\mathcal J(\mathbf Q_{n})^H \mathcal J(\mathbf Q_{n}) \right]^{-1}\mathcal G (\widehat{\boldsymbol \Gamma}_{n+1}).
\end{equation}
We propose to solve the denoising problem approximately.  Assuming $\lambda <<\beta$ and applying first-order Taylor approximation, we obtain an approximate solution for $\widehat{\mathbf Z}$ as
\begin{equation}
\label{denoiser_gen}
\widehat{\mathbf Z}_{n+1} \approx \underbrace{\left[\mathbf I ~-~\overbrace{\frac{\lambda}{\beta}~\mathcal J(\mathbf Q_n)^H \mathcal J(\mathbf Q_n)}^{\mathcal R_n} \right]}_{\textbf{$\mathcal L_n$}}\mathcal G (\widehat{\boldsymbol \Gamma}_{n+1}).
\end{equation}

\begin{figure}
\subfigure[Linear Residual Convolutional Block]{\includegraphics[width=0.5\textwidth,keepaspectratio=true,trim={5cm 7cm 5cm 6.8cm},clip]{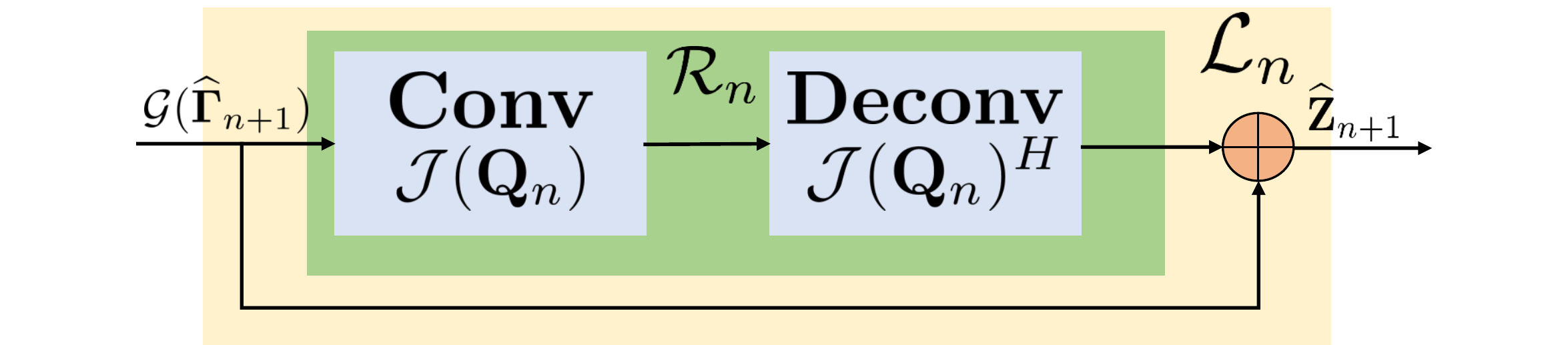}}\vspace{-1.1em}
\subfigure[Iterative Algorithm]{\includegraphics[width=0.48\textwidth,keepaspectratio=true,trim={4.5cm 6cm 4.2cm 5.3cm},clip]{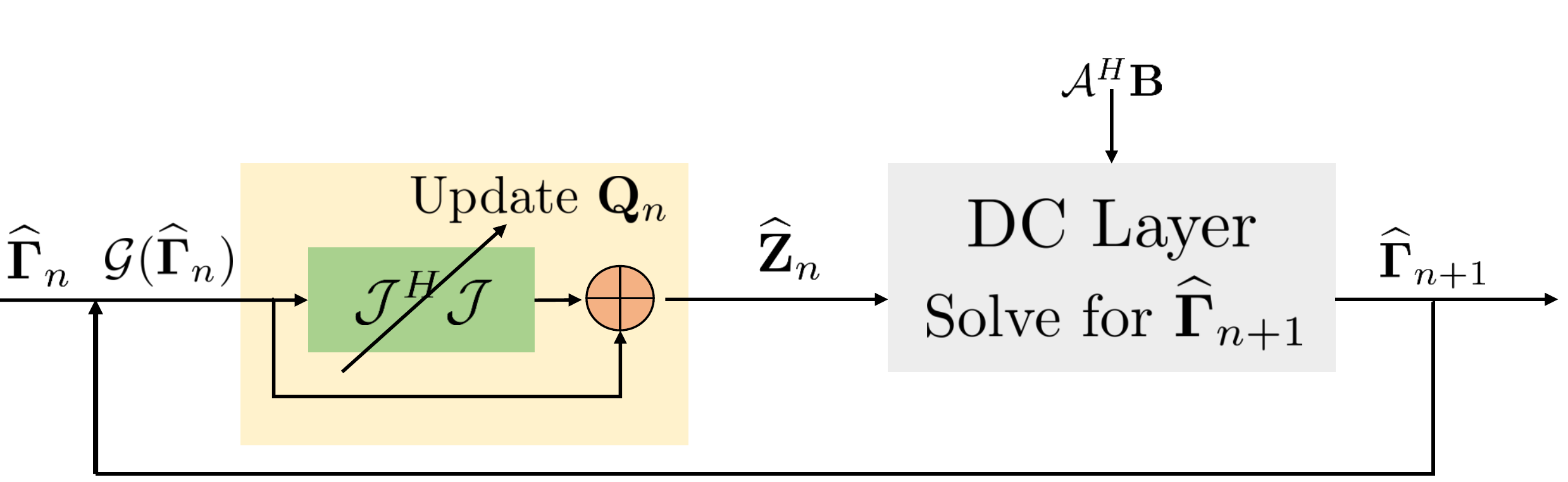}}
\caption{Illustration of the network structure of the IRLS algorithm used in structured low-rank algorithms: (a) shows the linear residual convolutional-deconvolutional block, which projects the signal at the $n^{\rm th}$ iteration to the signal subspace; (b) illustrates the network structure of the SLR algorithm, which alternates between the projection and the data consistency block.}\vspace{-1em}
	\label{irlsfig}
\end{figure}

As discussed before, $\mathcal J(\mathbf Q)$ denotes a MIMO or SIMO filterbank, depending on the nature of lifting. The term $\mathcal J(\mathbf Q)^H$ denotes convolution with a flipped version of $\mathbf Q$,  often referred to as the deconvolution layer in DL literature. $\mathcal R_n$ is a filterbank that projects the signal to the null space, thus killing or annihilating the signal and preserving the noise terms. Thus, the linear operator $\mathcal L_n$ is a residual block, which removes the alias or noise terms from the input signal, thus essentially denoising the signal (see Fig. \ref{irlsfig}). 

 Note that the filterbank $\mathbf Q_n$ has a subscript $n$ since it is updated at each iteration. The joint estimation of $\mathbf Q_n$ and reconstruction $\widehat{\boldsymbol{\Gamma}}_n$ results in high computational complexity. On the other hand, calibration-based methods pre-estimate $\mathbf Q$ and hence the residual filterbank $\mathcal L$, thus resulting in significantly reduced computational complexity. 
\subsection{SLR-inspired Model-based k-space DL}
\label{slrdeep}
The main disadvantage of the IRLS strategy discussed above is the high computational complexity. Specifically, this iterative approach requires an singular value decomposition (SVD) at each iteration, and thus results in a computationally expensive algorithm. To improve the computational efficiency, we propose to pre-learn a non-linear CNN annihilation filterbank $\mathcal N_{\rm k}$ from exemplar data. The subscript $\rm k$ indicates that the network performs convolutions in k-space.  
We pose a reconstruction similar to \eqref{lp}: 
\begin{equation}
\label{deep}
\arg \min_{{\boldsymbol \Gamma}} \|\mathcal{A}({\boldsymbol \Gamma})-\mathbf B\|_2^2 + \lambda_1 \|
\underbrace{\left(\mathcal I -\mathcal{D}_{\rm k}\right)}_{\mathcal{N}_{\rm k}}(\mathcal G(\widehat{\boldsymbol \Gamma}) )\|_2^2.
\end{equation}
Here, $\mathcal N_{\rm k}$ is a CNN that kills or annihilates the signal while preserving the noise or alias terms, which is conceptually similar to $\mathcal R_n$ in \eqref{denoiser_gen}. Thus, the operator $\mathcal D_{\rm k} = \mathcal I - \mathcal N_{\rm k}$ can be viewed as a denoiser similar to $\mathcal L_n$ in \eqref{denoiser_gen}. 

We propose to pre-learn the parameters of the network from exemplar data. Unlike calibrated schemes that learn a small linear network from a small subset of Fourier data (calibration region), the CNN parameters are learned from several fully sampled exemplar datasets. This approach enables us to learn a larger CNN, which can generalize to other datasets. We hypothesize that this pre-learned non-linear network can behave like a linear projection for each dataset, thereby facilitating their recovery from under-sampled data. Since the parameters of the network do not need to be self-learned, this approach is significantly faster than uncalibrated SLR approaches.

\begin{figure}[t!]
	\centering
	\subfigure[Residual CNN]{\includegraphics[width=0.4\textwidth,keepaspectratio=true,trim={6cm 5cm 6cm 5cm},clip]{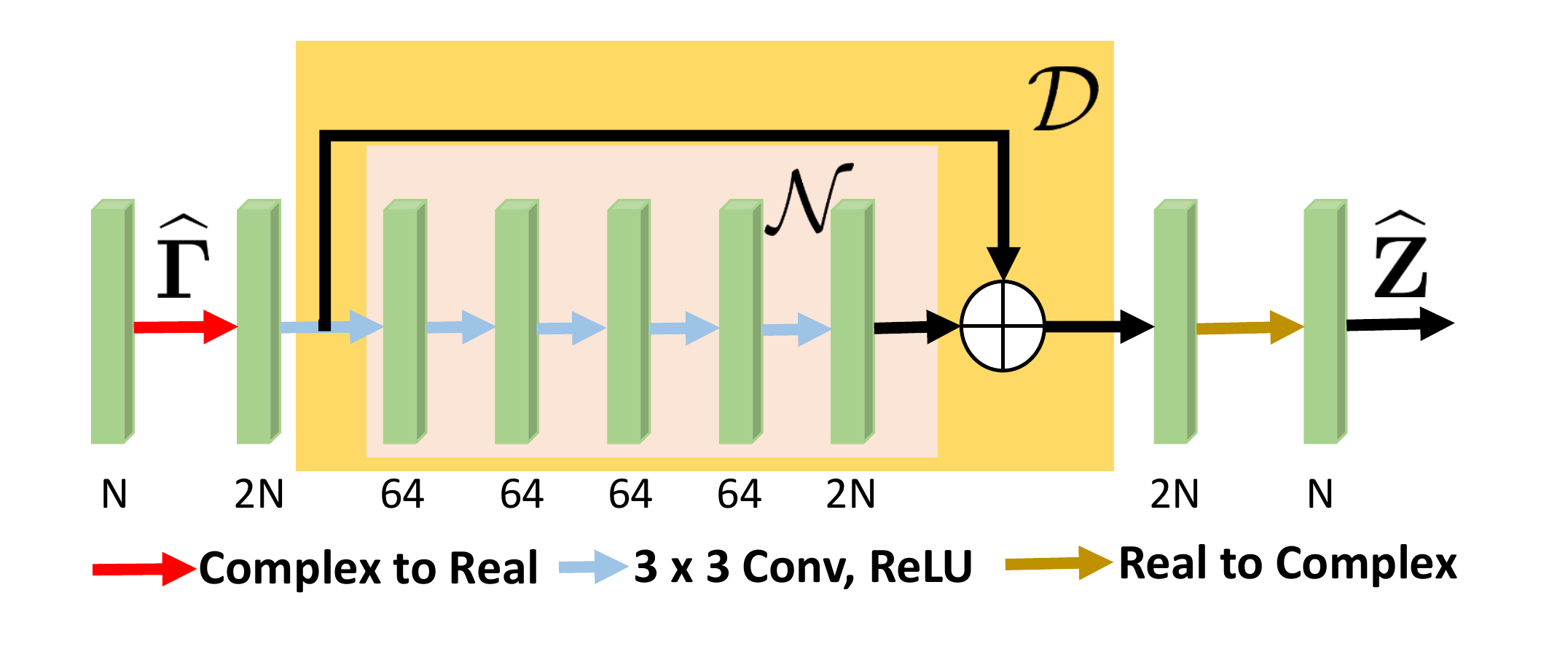}}\vspace{-1em}
	\subfigure[Proposed iterative algorithm]{\includegraphics[width=0.48\textwidth,keepaspectratio=true,trim={4.5cm 6cm 4.2cm 5cm},clip]{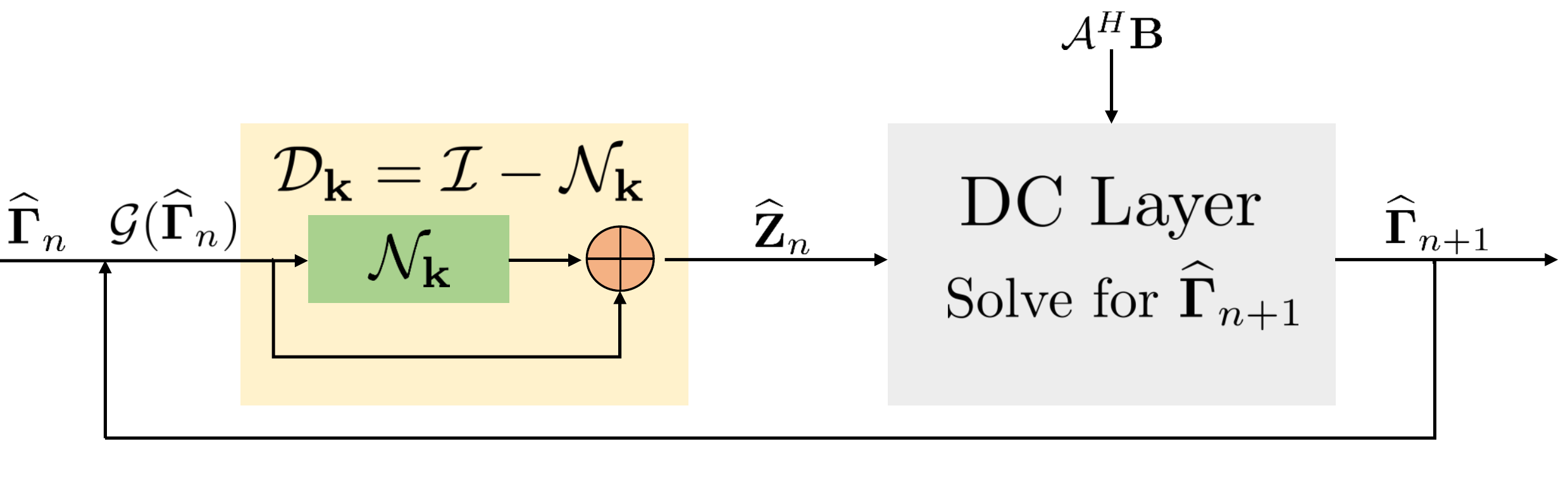}}
	\caption{Network structure of the proposed recursive CNN in k-space, described  in Section \ref{slrdeep}. The main difference of the proposed scheme with the approach in Fig. \ref{irlsfig} is the use of the deep residual CNN in (b), instead of the linear convolution-deconvolution block in Fig. \ref{irlsfig}.(a).}
	\label{fig:kspn}\vspace{-1em}
\end{figure}

We use an alternating minimization strategy similar to \eqref{img_update} and \eqref{z_update} to minimize \eqref{deep}. The resulting algorithm translates to a recursive network, which alternates between the denoising network $\mathcal D_{\rm k}$, which removes the noise and alias terms, and data consistency (DC) blocks: 
\begin{eqnarray}
\widehat{\mathbf Z}_n &=& \mathcal D_{\rm k}(\mathcal G(\widehat{\boldsymbol \Gamma}_n))\\
\widehat{\boldsymbol \Gamma}_{n+1} &=& (\mathcal A^H  \mathcal A + \lambda_1 \mathcal G^H \mathcal G)^{-1}(\mathcal A^H \mathbf B + \lambda_1 \mathcal G^H \widehat{\mathbf Z}_n)
\end{eqnarray} 
Similar to \cite{aggarwal2018modl}, we consider $K$ iterations of the above algorithm and unroll the above iterative scheme to obtain a deep network. The unrolled network consists of $K$ number of repetitions of both $\mathcal D_{\rm k}$ and DC blocks with parameters of $\mathcal D_{\rm k}$ being shared across iterations. At each iteration, the \emph{noisy} input $\mathcal G \widehat{\boldsymbol \Gamma}_n$ is projected to the signal subspace and hence denoised. The output of $\mathcal D_{\rm k}$ is given by $\mathcal D_{\rm k}(\mathcal G(\widehat{\boldsymbol \Gamma}_n))=\mathcal G(\widehat{\boldsymbol \Gamma}_n) - \mathcal N_{\rm k}(\mathcal G(\widehat{\boldsymbol \Gamma}_n))$.  The output is then fed into the DC block as shown in Fig \ref{fig:kspn}.  
As discussed previously, this iterative algorithm is similar to an alternating scheme to solve \eqref{lp}, with the distinction that the linear convolution-deconvolution block is replaced by a non-linear CNN. Unlike the setting in \eqref{lp}, where the filter parameters are learned from the calibration data of each dataset, we propose to pre-learn a CNN from exemplary data. 
\subsection{Hybrid Regularized DL}
\begin{figure}
	\includegraphics[width=0.48\textwidth,keepaspectratio=true,trim={4cm 7cm 3.5cm 5.7cm},clip]{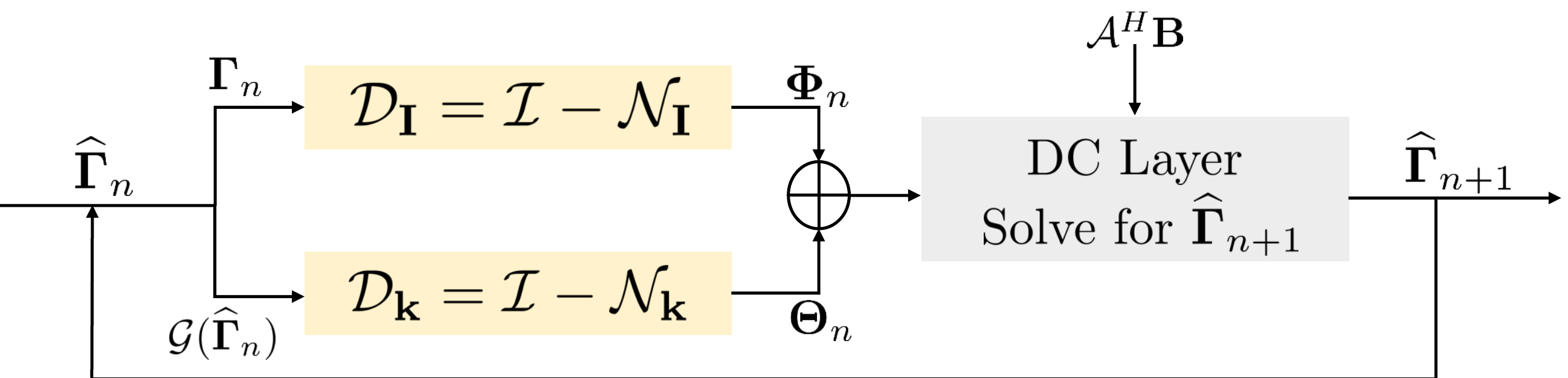}
	\caption{Hybrid network: It consists of two identically structured residual CNNs $\mathcal D_k$, $\mathcal D_{\rm I}$ for k-space and image domain learning, respectively. The $\mathcal D_{\rm I}$ block learns redundancies in patches, while $\mathcal D_k$ block exploits k-space annihilation relations. The $\mathcal D_{\rm I}$ block does an inverse fast Fourier transform (IFFT) on input k-space ${\boldsymbol \Gamma}$ and passes it to the residual CNN. The residual image output is transformed back to k-space ${\boldsymbol \Gamma}$ by a fast Fourier transform (FFT). The output of $\mathcal D_{\mathbf k}$ and $\mathcal D_{\mathbf I}$ at the $\rm n^{th}$ iteration are denoted by $\boldsymbol \Theta_n$ and $\boldsymbol \Phi_n$ according to \eqref{ksp_proj_hybrid}, \eqref{img_proj_hybrid} and \eqref{dchybrid}.  The parameters are not shared between $\mathcal D_k$ and $\mathcal D_{\rm I}$. Both $\mathcal D_k$ and $\mathcal D_{\rm I}$ in hybrid Deep-SLR have half the number of feature maps per layer compared to $\mathcal D_k$ in k-space Deep-SLR to keep number of trainable parameters the same in both networks for fair comparison. The network parameters are shared across iterations similar to the MoDL \cite{aggarwal2018modl} framework.}
	\label{fig:hybn}
\end{figure}

The SLR methods exploit the redundancies in k-space resulting from specific structures in the signal. However, the image patches in MR images often exhibit extensive redundancy, which is exploited in our MoDL scheme \cite{aggarwal2018modl} as well as other image domain methods \cite{lee2018deep,schlemper2017deep,hammernik2018learning}. These priors are complementary to the SLR priors discussed in the previous section. We propose to modify the cost function in \eqref{deep} as
\begin{equation}
\label{deep_hybrid}
\arg \min_{\boldsymbol \Gamma} \|\mathcal{A}({\boldsymbol \Gamma})-\mathbf B\|_2^2 + \lambda_1 \|\mathcal{N}_{\rm k}(\mathcal G(\widehat{\boldsymbol \Gamma})) \|_2^2 + \lambda_2 \|\mathcal{N}_I ({\boldsymbol \Gamma}) \|_2^2.
\end{equation}  
Here, $\mathcal N_{\mathbf I} $ and $\mathcal N_{\mathbf k} $ are two residual CNNs. The alternating minimization of this scheme results in the following steps:
\begin{eqnarray}
\label{ksp_proj_hybrid}
\boldsymbol \Theta_n &=& \mathcal D_{\rm k}(\mathcal G (\widehat{\boldsymbol \Gamma}_n))\\
\label{img_proj_hybrid}
\boldsymbol \Phi_n &=& \mathcal D_{\rm I}(\boldsymbol \Gamma_n)\\\nonumber
\label{dchybrid}
\widehat{\boldsymbol \Gamma}_{n+1} &=& (\mathcal A^H\mathcal A + \lambda_1 \mathcal G^H \mathcal G + \lambda_2 \mathcal I)^{-1}(\mathcal A^H \mathbf B \\& &+ \lambda_1 \mathcal G^H \boldsymbol \Theta_n  + \lambda_2\boldsymbol \Phi_n)
\end{eqnarray} 
as shown in Fig \ref{fig:hybn}. The $\mathcal D_{\rm k}$ relies on annihilation relations in k-space, while $\mathcal D_{\rm I}$ exploits the image domain priors. We propose to learn the parameters of the CNNs $\mathcal D_{\rm k}$ and $\mathcal D_{\rm I}$ using exemplary data.
\subsection{Special Cases}
We show applications of our proposed methods in both single-channel and multi-channel settings, and show that the image sub-problem can be solved analytically in both cases. This approach will accelerate the training and testing procedures.
\subsubsection{Piecewise Constant Image Structure}
The GIRAF \cite{ongie2017fast} algorithm is an SLR scheme that exploits the piecewise constant nature of images, as described in Section \ref{girafsec}. Here, the operator $\mathcal G(\widehat{\boldsymbol{\gamma}}) = \widehat{\nabla \boldsymbol \gamma}$ as defined in \eqref{nabla_def}. In this case, we have   
\begin{eqnarray}
\mathcal G^H\left(\begin{bmatrix}
\widehat{\mathbf z}_1\\\widehat{\mathbf z}_2
\end{bmatrix}\right)[\mathbf k] &=& -(j2\pi k_x \widehat{\mathbf z}_1[\mathbf k] +j2\pi k_y \widehat{\mathbf z}_2[\mathbf k] )\\
\mathcal G^H\mathcal G\left(\widehat{\boldsymbol{\gamma}}\right)[\mathbf k] &=& 4\pi^2 (k_x^2+k_y^2) ~\widehat{\boldsymbol \gamma}[\mathbf k]
\end{eqnarray}
Note that the matrix $(\mathcal A^H  \mathcal A + \lambda_1 \mathcal G^H \mathcal G + \lambda_2 \mathcal I)$ in \eqref{dchybrid} can be viewed as a weighting operator in the Fourier domain in the single-channel setting. We can thus solve \eqref{dchybrid} analytically.
\subsubsection{Parallel MRI Acquisition}
In a parallel MRI setting,  $\mathcal G=\mathcal I$ and hence the data consistency term simplifies to $ (\mathcal A^H  \mathcal A + \left(\lambda_1  + \lambda_2\right) \mathcal I)^{-1}(\mathcal A^H \mathbf B + \lambda_1 \boldsymbol \Theta_n + \lambda_2 \boldsymbol \Phi_n)$. The term $(\mathcal A^H\mathcal A + \left(\lambda_1  + \lambda_2\right) \mathcal I)$ is separable across the channels. Hence, one can independently solve for each channel of $\widehat{\boldsymbol \Gamma}_{n+1}$ in the Fourier domain in an analytical fashion.

\section{Implementation Details}
\subsection{Datasets}
\label{data_acq}
The datasets used for single channel experiments were multi-coil k-space of knee from (\url{www.mridata.org}) and multi-coil brain from the Calgary-Campinas Public (CCP) dataset \cite{souza2018open} in (\url{https://sites.google.com/view/calgary-campinas-dataset}). The CCP consists of 12-channel T1-weighted brain MR datasets acquired on a 3T scanner. It is a 3D acquisition that allows undersampling along two directions (phase and slice encoding). We used the single channel complex valued images provided by the organizers, which were generated by multi-channel coil combination. The Fourier Transform (FT) was applied to obtain k-space samples from the coil combined images. Since the frequency encoding dimension is fully sampled, we performed an IFFT along this dimension and considered the recovery of each 2D slice independently. We chose twenty subjects for training, five for validation, and ten for testing. The other dataset consisting of multi-channel knee data of twenty subjects was acquired with a 3D fast spin echo (FSE) sequence on a 3T scanner. The parameters set for the scan were: repetition time TR = 1550 ms, echo time TE = 25 ms, and a flip angle of $90^{\circ}$. There are 256 sagittal slices and 320 coronal slices per subject with matrix sizes of 320 x 320 and 320 x 256, respectively, at a slice thickness of 0.5 mm. A coil combination of the 8-channel knee k-space data was performed using principal component analysis (PCA). Specifically, we performed a PCA along the coil dimension and picked the first component along the coil dimension as the single-channel complex image. This coil compression preserved on average about 90\% of the energy from multi-channel data. Fifteen subjects were used for training, two for validation, and the remaining three for testing. 

A set of experiments were done to study the workings of the k-space Deep-SLR scheme for single-channel MRI recovery. The NIfTI formatted T2-weighted brain datasets from the Human Connectome Project (HCP) \cite{van2012human} were used. The T2-weighted brain images were acquired by a Siemens 3T MR scanner using a 3D Cartesian spin-echo sequence. The TR and TE parameters were 3200 ms and 565 ms, respectively, while the matrix size was 320 x 256 with a field of view (FOV) of 224 x 224 mm$^{2}$. 

Parallel MRI experiments were performed on multi-channel brain and knee datasets. The knee dataset \cite{hammernik2018learning} is a multi-slice 2D dataset consisting of 15-channel slices from 20 subjects with roughly 40 slices per subject. The slices are of dimension 640 x 368 x 15. Twelve subjects were used for training, one for validation, and the remaining seven for testing. The data was under-sampled by varying density along the phase encodes. Brain MRI was collected from nine subjects at University of Iowa Hospitals and Clinics using a 3D T2 CUBE sequence with Cartesian readouts using a 12-channel head coil. There are 140 3D slices per subject with dimensions 12 x 256 x 232. We used five subjects for training, one for validation, and the remaining three for testing. 

In both cases, the fully sampled complex k-space data was under-sampled and used for training. The complex image obtained by evaluating the IFFT of the individual coil data was used as ground truth in training and testing.

\begin{figure*}
	\centering
	\includegraphics[width=\textwidth,keepaspectratio=true,trim={1.9cm 9.1cm 1.2cm 9.5cm},clip]{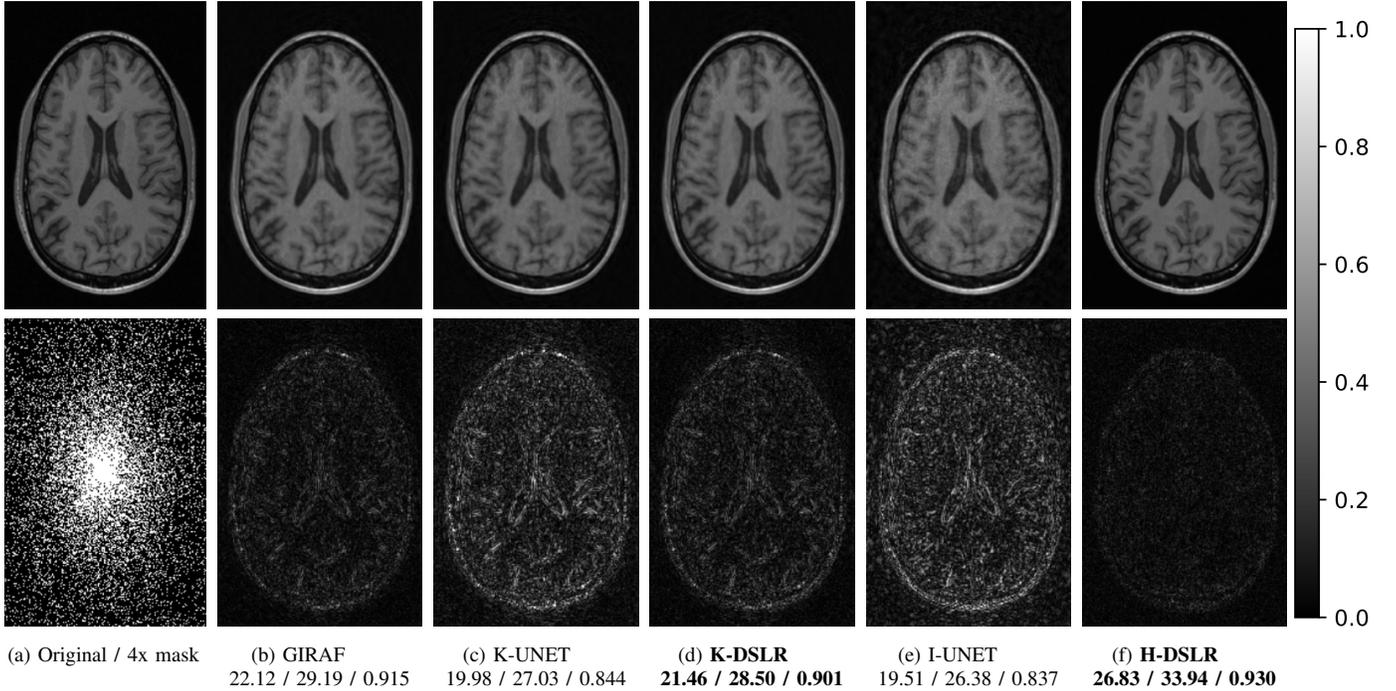}
	\caption{Reconstruction results of 4x accelerated single-channel brain data. SNR (dB)/PSNR (dB)/SSIM values are reported for each case. The data was under-sampled using a cartesian 2D non-uniform variable-density mask. The top row shows reconstructions (magnitude images), while the bottom row shows corresponding error images. The additional image domain prior in \textbf{H-DSLR} ensures significant improvement in performance over other schemes.}
	\label{fig:sc_brain_rec}
\end{figure*}
\subsection{Quality Evaluation Metric}
We quantitatively evaluate the recovered images in terms of signal-to-noise ratio (SNR) and structural similarity (SSIM) index. The SNR of an image is computed as $
\mathbf{SNR} = 20 \cdot \log_{10} \left(\frac{\|\mathbf x_{\rm rec}\|_2}{\|\mathbf x_{\rm org}-\mathbf x_{\rm rec}\|_2}\right)$, where $\mathbf x_{\rm org}$ and $ \mathbf x_{\rm rec}$ are original ground truth and reconstructed images, respectively.
\subsection{Architecture of the CNNs}
The modular nature of the proposed scheme allows us to use any residual CNN architecture to define the prior. A key difference with the approach in Fig. \ref{irlsfig} is that the CNN parameters are fixed and do not change with iterations as in Fig. \ref{irlsfig}.(b). The pre-learning of the CNN parameters using exemplar data allows us to significantly reduce the number of alternating steps compared to the self-learning strategy in Fig. \ref{irlsfig}. 
Image domain CNN $\mathcal N_{\rm I}$ is structurally identical to the Fourier domain CNN $\mathcal N_{\rm k}$, with an equal number of parameters. The residual block $\mathcal D_{\rm I}$ performs an IFFT that feeds spatial domain input $\boldsymbol \Gamma$ to the CNN $\mathcal N_{\rm I}$ and transforms the residual output back to k-space by a fast Fourier transform (FFT) operation. For implementation purposes, we split the real and imaginary parts of the input k-space data into real and imaginary components, which are fed as two channels. The two output channels are combined to recreate the complex output k-space data. The models were implemented  
\subsubsection{Single-channel Case} We use a residual UNET as $\mathcal N_{\rm k}$ in the single-channel setting for the proposed k-space Deep-SLR (K-DSLR) scheme. We use its modified version of UNET with only 12 layers (two pooling and unpooling operations). The number of filters per layer grows from 64 to a maximum of 256. The UNET operates on single-channel Fourier data ($M=1$). For the proposed hybrid scheme H-DSLR, the number of parameters in both the UNETs were halved layer by layer to keep the total similar to the K-DSLR network for fair comparisons. 
\subsubsection{Parallel MRI (Multi-channel Case)}A residual five-layer MIMO CNN $\mathcal N_{\rm k}$ as shown in Fig. \ref{fig:kspn}.(b) is used as the k-space network in K-DSLR. The input and output channels of the network are adjusted according to the dataset. For example, $M = 12$ and $M = 15$ channels are set for multi-channel brain and knee data, respectively. Each convolution layer consists of 64 3 x 3 filters, followed by ReLU non-linearity. The number of filters per layer was halved to 32 for both the CNNs in H-DSLR compared to 64 in K-DSLR for fair comparison. 

We trained the unrolled recursive network for different iterations of $K$. $K=10$ was found to be the best-performing model on test data for both cases and the performance saturated afterwards. We were constrained by 16 GB GPU memory, which restricted us from going beyond 15 iterations. The regularization parameters were fixed at $\lambda_1 =  \lambda_2 = 1$ for all the experiments. The weights were Xavier initialized and trained for 500 epochs with an Adam optimizer to reduce the mean square error (MSE) at a learning rate of $10^{-4}$. All the DL models were implemented using Tensorflow version 1.15. The proposed $K=10$ iteration models for single and multi-channel cases took 5 and 10 hours for training respectively. The source code for the proposed H-DSLR scheme on multi-channel MRI datasets can be viewed and downloaded from the github link: \url{https://github.com/anikpram/Deep-SLR} .    

\subsection{State-of-the-art Methods for Comparison}
We compare our scheme for single-channel recovery against the SLR algorithm (GIRAF) \cite{ongie2017fast}, a k-space UNET (K-UNET) \cite{han2019k}, and an image domain UNET (I-UNET). The K-UNET is a direct DL approach with a 20-layer 2D UNET in k-space without a DC step. It accepts a real image formed by concatenation of real and imaginary parts of 2D complex k-space. The I-UNET is the spatial version of K-UNET where learning is performed in spatial domain. The I-UNET structure and its number of parameters are exactly the same as in K-UNET. These networks were trained and tested on single-channel knee and brain datasets described in Section \ref{data_acq}.

In the parallel MRI setting, we compare the proposed scheme with MoDL \cite{aggarwal2018modl}, K-UNET \cite{han2019k}, and the calibration-less parallel SLR algorithm, which motivated our proposed scheme. K-UNET is also a multi-channel calibration-less direct DL approach in k-space without a DC step \cite{han2019k}. Its structure is similar to single-channel K-UNET, with the only difference being the multi-channel input and output. MoDL \cite{aggarwal2018modl} is a pre-calibrated approach that uses coil sensitivity information and spatial domain regularization. The coil sensitivities for MoDL were estimated using ESPIRiT \cite{uecker2014espirit}. All the parallel MRI methods were evaluated on the brain and knee datasets mentioned in Section \ref{data_acq}.
\section{Experiments and Results}
Experiments were done on multiple datasets for both single-channel sparse MRI and parallel MRI recovery. Additional experiments were also done on diffusion MRI recovery and are discussed in the supplementary material.
\subsection{Single-channel Signal Recovery}
Comparisons of the proposed single-channel schemes against state-of-the-art methods are shown in Fig. \ref{fig:sc_brain_rec} for the CCP dataset described in Section \ref{data_acq}. We observe that the k-space Deep-SLR (K-DSLR) approach in Fig. \ref{fig:sc_brain_rec}.(d) provides results that are comparable to the model-based GIRAF \cite{ongie2017fast} method in Fig. \ref{fig:sc_brain_rec}.(b). By contrast, the direct inversion based I-UNET and K-UNET provides lower performance, even though the number of trainable parameters are larger. Of these, the K-UNET provides slighly lower errors. The improved performance of K-DSLR over K-UNET may be attributed to the model-based approach, which repeatedly enforces DC. Fig. \ref{fig:sc_brain_rec}.(f) corresponds to H-DSLR, which uses both k-space and image domain priors. The H-DSLR scheme significantly reduces errors. The number of parameters in this model is similar to the one in Fig. \ref{fig:sc_brain_rec}.(d) since the number of output channels of each intermediate layer is halved. However, the addition of the complementary prior significantly reduces the errors. A similar set of experiments were also performed on the single-channel coronal and sagittal view of knee images described in Section \ref{data_acq}. The comparisons are shown and discussed in the supplementary paper. The quantitative results of all the experiments are recorded in Table S1 in the supplementary section. 
\begin{figure}[b!]
	\centering
	\includegraphics[width=0.5\textwidth,keepaspectratio=true,trim={1.8cm 10.9cm 11.5cm 11.4cm},clip]{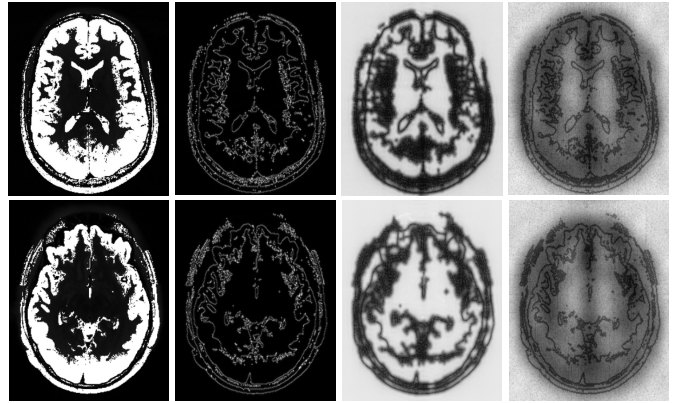}
	\caption{Illustration of the non-linear and linear annihilation operators. Piecewise constant images and their gradients are shown in (a) and (b), respectively. The non-linear block $\mathcal N_{\rm k}$ behaves like a linear projector to the null space for each image.  Pseudo-random perturbations of small magnitude are added to the gradients of the image and fed to $\mathcal N_{\rm k}$. The SOS of the output perturbations are shown in (d). The SOS function on $\mathcal N_{\rm k}$ closely mimics the linear operator $\mathcal R$ in (c). Specifically, it annihilates or kills the gradient components close to the edge locations while preserving the noise components far from the edges. We show more results on different slices in the supplementary material (see Fig. S1).}
	\label{fig:psf}
\end{figure}   
\begin{figure*}[t!]
	\centering
	\includegraphics[width=\textwidth,keepaspectratio=true,trim={1.8cm 9.05cm 0.6cm 9.8cm},clip]{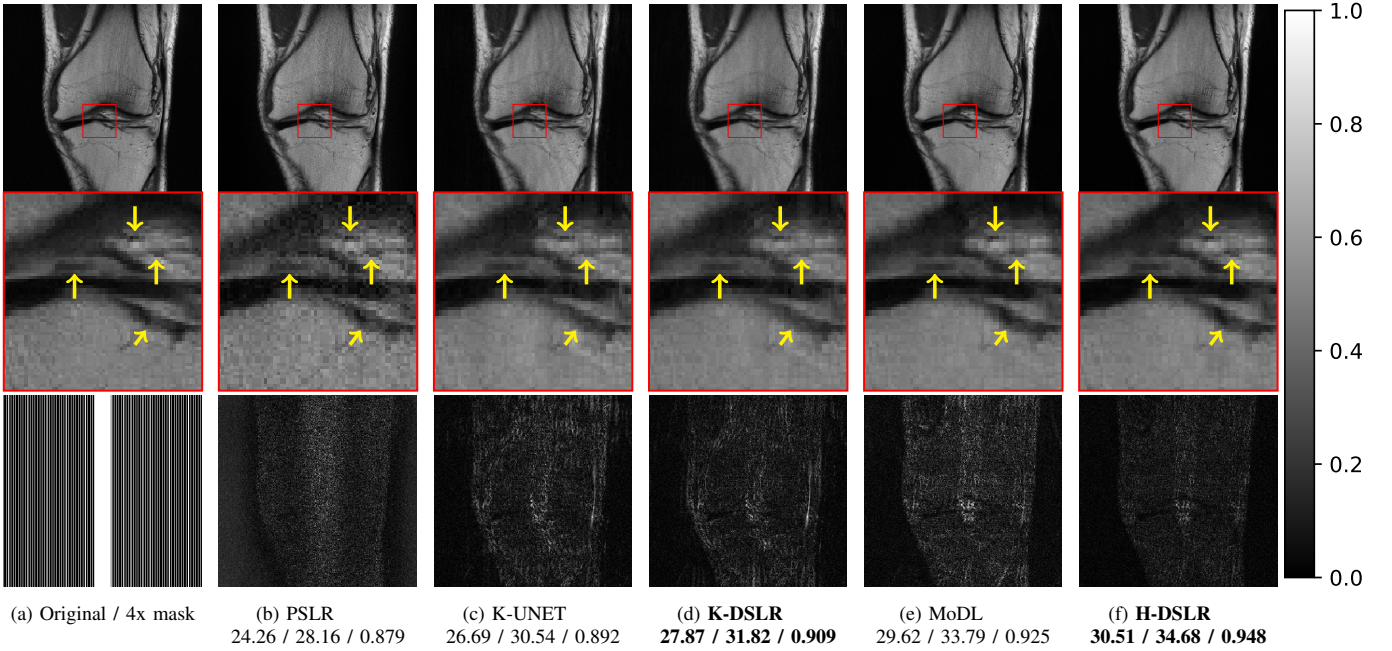}
	\caption{Reconstruction results of 4x accelerated 15-channel knee data. A 2D Cartesian structured under-sampling along phase encodes was done. The top row displays reconstructions (SOS), and the bottom row shows corresponding error images. The yellow arrows in the zoomed cartilage region show minute details better preserved by the proposed scheme over other state-of-the-art methods. The numbers are SNR reported in dB. The k-space Deep-SLR scheme \textbf{K-DSLR} yields comparable results to the parallel SLR scheme. The addition of the image domain prior further improves performance. We show H-DSLR reconstructions of different slices with different accelerations factors in the supplementary material (see Fig. S5).}
	\vspace{-1em}
	\label{fig:mc_knee_florian}
\end{figure*}
\subsection{Annihilation Operators on Piecewise Constant Images}
\label{hypo_test}
Single-channel Deep-SLR scheme is used to study its
inner workings and its similarity to classical SLR methods in Fig. \ref{fig:psf}. Note that k-space SLR methods for single-channel MRI schemes \cite{ongie2015super,ongie2017convex} learn linear annihilation relations in k-space. As shown in the literature \cite{ongie2015super,ongie2017convex}, the SLR penalty in \eqref{lp} is a weighted $\ell_2$ norm of the gradients of the image, where the weights correspond to the sum of squares (SOS) of the estimated null space filters. Specifically, the linear annihilation operator has several linearly independent null space vectors; the sum of squares of the IFFT of the null space vectors yield zeros in the location of the gradients in the single-channel setting as shown in the literature \cite{ongie2017fast}. The SLR scheme estimates annihilation relations from under-sampled data using an optimization strategy. By contrast,  the proposed scheme learns to estimate the annihilation relations from under-sampled measurements based on its training on exemplar data.

The solutions provided by the unconstrained setting considered in this paper and \cite{aggarwal2018modl} are similar to constrained setting in \cite{tamir2019unsupervised}, where the formulation in \eqref{deep} is replaced by 
\begin{equation}\label{unconstrained}
\boldsymbol{\gamma}^* = \arg \min \|\mathcal N_{\rm k}(\mathcal G(\widehat{\boldsymbol \gamma}))\|^2 ~~\mbox{such that}~~\|\mathcal A(\boldsymbol \gamma)-\mathbf b\|^2 \leq \sigma^2, 
\end{equation}
where $\sigma$ is the noise variance; see \cite{tamir2019unsupervised} for detailed performance comparisons of the constrained and unconstrained formulations. In this case, the data consistency layer specified by \eqref{dchybrid} gets modified as
\begin{equation}\label{dclayer}
\boldsymbol{\gamma}_{n+1} = \arg \min \|\mathcal G(\widehat{\boldsymbol \gamma}) - \mathbf{\widehat z}_n\|^2 ~~\mbox{such that} ~~\|\mathcal A(\boldsymbol \gamma)-\mathbf b\|^2 \leq \sigma^2.
\end{equation}
When $\mathcal A$ satisfies the restricted isometry conditions \cite{candes2008restricted}, then $
\epsilon \|\boldsymbol\gamma-\boldsymbol\gamma^*\|^2 \leq \mathcal A(\boldsymbol\gamma-\boldsymbol\gamma^*)\leq\delta \|\boldsymbol\gamma-\boldsymbol\gamma^*\|^2$, where $\epsilon$ and $\delta$ are the restricted isometry property (RIP) constants. Thus,
\begin{equation}
\|\boldsymbol \gamma_{n+1} -\boldsymbol{\gamma}^*\|^2\leq \frac{\sigma^2}{\epsilon},
\end{equation}
where $\boldsymbol \gamma^*$ is the true solution. This relation implies that at each iteration $n$, the input to the network $\boldsymbol \gamma_{n}$ is within a $\sigma^2/\epsilon$ ball of the true solution $\boldsymbol \gamma^*$.

We note that an arbitrary non-linear function can be approximated by its first-order Taylor series representation in a small neighborhood. Our hypothesis is that the first order Taylor series approximation of the non-linear annihilation block $\mathcal N_{\rm k}(\mathcal G(\widehat{\boldsymbol \gamma}))$ within the $\sigma^2/\epsilon$ ball around $\boldsymbol \gamma^*$ closely matches the linear annihilation relations in SLR schemes. Specifically, the annihilation filters would kill the high gradients, while preserving the noise. The use of this annihilation filterbank within the residual block, results in preserving the true signal while suppressing the noise-like perturbations.

In order to test this hypothesis, small random perturbations of variance $\sigma = 0.01$ are added to a given image $\boldsymbol \gamma^*$ and the corresponding output perturbations are analyzed; the sum of squares of the corresponding perturbations is an indicator of the response of the annihilation operator. We consider a piecewise constant image in Fig. \ref{fig:psf}, which was derived from an image from the HCP dataset (described in Section \ref{data_acq}), by thresholding. The CNN network with the same architecture as above (12-layer UNET as $\mathcal N_{\rm k}$) are trained using piecewise constant brain images from 10 training subjects, also obtained by thresholding the HCP data. Following training, random perturbations are added to a new dataset and its k-space data is passed through the network. \\ The sum of squares of the IFFT of the outputs for 1000 realizations are evaluated, which are shown in Fig. \ref{fig:psf}.(d). Note that the zeros of the SOS output function closely mimics the SOS function in Fig. \ref{fig:psf}.(c). This behaviour is observed across a wide variety of testing slices unseen by the trained network, as shown in Fig. \ref{fig:psf} and also in Fig. S1 in the supplemental document, which justifies its generalizability. The experiment strengthens our hypothesis that the proposed network behaves as a linear projector similar to classical SLR schemes for each image $\boldsymbol \gamma^*$. While similar results are observed for natural images, it is difficult to visualize this due to the large dynamic range. We note that without additional constraints on the weights, one cannot guarantee that the Lipschitz constants of the network is bounded for all inputs, including adversarial perturbations.

\subsection{Parallel MRI Recovery}
\begin{figure*}[t!]
	\centering
	\includegraphics[width=\textwidth,keepaspectratio=true,trim={1.8cm 9.8cm 1.1cm 10.25cm},clip]{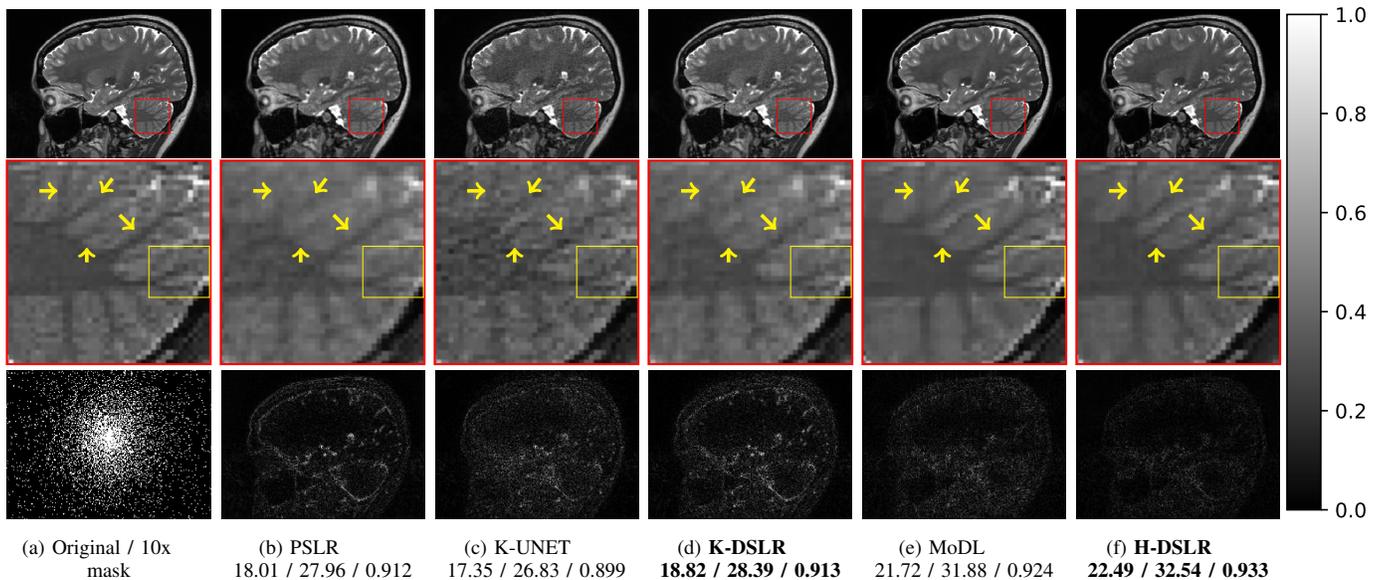}
	\caption{Reconstruction results of 10x accelerated 12-channel brain data. SNR (dB)/PSNR (dB)/SSIM values are reported for each case. The under-sampling pattern was chosen to be a 2D Cartesian non-uniform variable density. The top row images are reconstructions (SOS), while the bottom row shows corresponding error images. The yellow arrows in the zoomed cerebellum region show minute details better preserved by the proposed scheme than by other state-of-the-art methods. The \textbf{K-DSLR} scheme has errors of lower magnitude than the calibration-less k-space methods PSLR and K-UNET. The proposed hybrid scheme \textbf{H-DSLR} performs comparably to the pre-calibrated approach MoDL. We show H-DSLR reconstructions of different slices with different accelerations factors in the supplementary material (see Fig. S4).}
	\vspace{-1.4em}
	\label{fig:mc_brain}
\end{figure*}

Proposed multi-channel schemes are compared against state-of-the-art calibration-less and calibrated schemes in Figures \ref{fig:mc_knee_florian} and \ref{fig:mc_brain}, and Table \ref{tab:comp_mc}. The methods have been tested on 420 3D brain slices collected from three subjects. The same set of methods have also been tested on approximately 300 (seven subjects) 3D knee slices. Similar to the single-channel case, the performance of the multi-channel K-DSLR is comparable to the parallel SLR (PSLR) scheme. The k-space network exhibits some residual aliasing in the knee example in Fig. \ref{fig:mc_knee_florian}.(d), which can be attributed to the highly structured/uniform nature of sampling. Note that the data was acquired with a calibration region, which the iterative PSLR scheme seems to have benefited from, even though we did not explicitly rely on a calibrated approach. The table reveals that the proposed H-DSLR outperformed the multi-channel PSLR and K-UNET \cite{han2019k} and that it is slightly better than the pre-calibrated approach MoDL \cite{aggarwal2018modl}. 
Note that MoDL is a calibrated scheme, which requires the explicit knowledge of the coil sensitivities. The coil sensitivities were estimated from the fully sampled region in the center of k-space for both knee (Fig. \ref{fig:mc_knee_florian}) and brain (Fig. \ref{fig:mc_brain}) experiments using ESPIRiT \cite{uecker2014espirit}. The calibration-less methods compared here (PSLR, the proposed method and K-UNET) perform an interpolation in k-space without explicit knowledge of the coil sensitivities. The addition of the image domain prior (H-DSLR in Fig. \ref{fig:mc_knee_florian}.(f)) is found to suppress the artifacts and provide reconstructions that are comparable to the MoDL scheme. The proposed Deep-SLR scheme facilitates the recovery of the images without the knowledge of the coil sensitivities. This approach thus eliminates the potential mismatch between the calibration scans for the estimation of the coil sensitivities and the main scan in approaches that rely on an extra calibration scans. By removing the need for an explicit calibration region, this approach enables higher acceleration factors. An additional study of the robustness of our proposed approach to acceleration factors for both knee and brain datasets is presented in the supplementary section of this paper. 

\begin{table}[b!]
	\fontsize{6}{8}
	\selectfont
	\centering
	\renewcommand{\arraystretch}{0.8}
	\begin{tabular}{|c|cc|c|}
		\hline
		\multicolumn{4}{|c|}{Signal-to-Noise Ratio (SNR)}\\ \hline
		\multicolumn{3}{|c|}{Brain} & \multicolumn{1}{|c|}{Knee} \\ \hline
		Acceleration  & \multicolumn{1}{c}{6x} & \multicolumn{1}{c}{10x} & \multicolumn{1}{|c|}{4x} \\
		Methods & SNR  & SNR & SNR  \\ \hline 
		PSLR &21.02 $\pm$ 2.33  &18.12 $\pm$ 2.58  & 24.26 $\pm$ 2.12  \\
		K-UNET &19.58 $\pm$ 2.01  &17.28 $\pm$ 1.98  & 26.81 $\pm$ 2.05 \\
		\textbf{K-DSLR} & 21.58 $\pm$ 1.74 & 18.71 $\pm$ 1.83 & 27.87 $\pm$ 1.36  \\
		MoDL &23.30 $\pm$ 1.53 &21.63 $\pm$ 1.62  & 29.77 $\pm$ 1.19 \\
		\textbf{H-DSLR} & 24.34 $\pm$ 1.15 & 22.20 $\pm$ 1.23  & 30.57 $\pm$ 0.96  \\ \hline
        
        \multicolumn{4}{|c|}{Peak Signal-to-Noise Ratio (PSNR)}\\ \hline
		\multicolumn{3}{|c|}{Brain} & \multicolumn{1}{|c|}{Knee} \\ \hline
		Acceleration  & \multicolumn{1}{c}{6x} & \multicolumn{1}{c}{10x} & \multicolumn{1}{|c|}{4x} \\
		Methods & PSNR  & PSNR & PSNR  \\ \hline 
		PSLR &31.17 $\pm$ 2.30  &28.21 $\pm$ 2.61  & 28.19 $\pm$ 2.03  \\
		K-UNET &29.49 $\pm$ 1.96  &27.14 $\pm$ 1.91  & 30.94 $\pm$ 2.14 \\
		\textbf{K-DSLR} & 31.66 $\pm$ 1.77 & 28.55 $\pm$ 1.84 & 31.67 $\pm$ 1.33  \\
		MoDL &33.43 $\pm$ 1.51 &31.53 $\pm$ 1.57  & 33.77 $\pm$ 1.19 \\
		\textbf{H-DSLR} & 34.46 $\pm$ 1.22 & 32.31 $\pm$ 1.23  & 34.69 $\pm$ 0.98  \\ \hline

        \multicolumn{4}{|c|}{Structural Similarity (SSIM)}\\\hline
		\multicolumn{3}{|c|}{Brain} & \multicolumn{1}{|c|}{Knee} \\ \hline
		Acceleration  & \multicolumn{1}{c}{6x} & \multicolumn{1}{c}{10x} & \multicolumn{1}{|c|}{4x} \\		
Methods &  SSIM  & SSIM  & SSIM \\ \hline 
		PSLR  &0.942 $\pm$ 0.035  &0.918 $\pm$ 0.041  &0.873 $\pm$ 0.027 \\
		K-UNET  &0.920 $\pm$ 0.029 &0.883 $\pm$ 0.030  &0.887 $\pm$ 0.021 \\
		\textbf{K-DSLR}  & 0.938 $\pm$ 0.018 & 0.913 $\pm$ 0.023  & 0.904 $\pm$ 0.011  \\
		MoDL  &0.951 $\pm$ 0.020 &0.921 $\pm$ 0.026  &0.928 $\pm$ 0.015 \\
		\textbf{H-DSLR}  & 0.958 $\pm$ 0.011 & 0.935 $\pm$ 0.013  & 0.944 $\pm$ 0.008  \\ \hline		
	\end{tabular}
	\vspace{1em}
	\caption{Quantitative comparison of PSLR, MoDL, proposed, and UNET reconstructions in terms of SNR (dB), PSNR (dB) and SSIM. The bold-faced methods are the proposed ones.}
	\label{tab:comp_mc}	 
\vspace{-2em}
\end{table}

\subsection{Benefits over Calibrated Approaches}
\begin{figure*}[t!]\vspace{1.5em}
	\centering
	\includegraphics[width=0.8\textwidth,keepaspectratio=true,trim={1.7cm 9.6cm 5.7cm 10cm},clip]{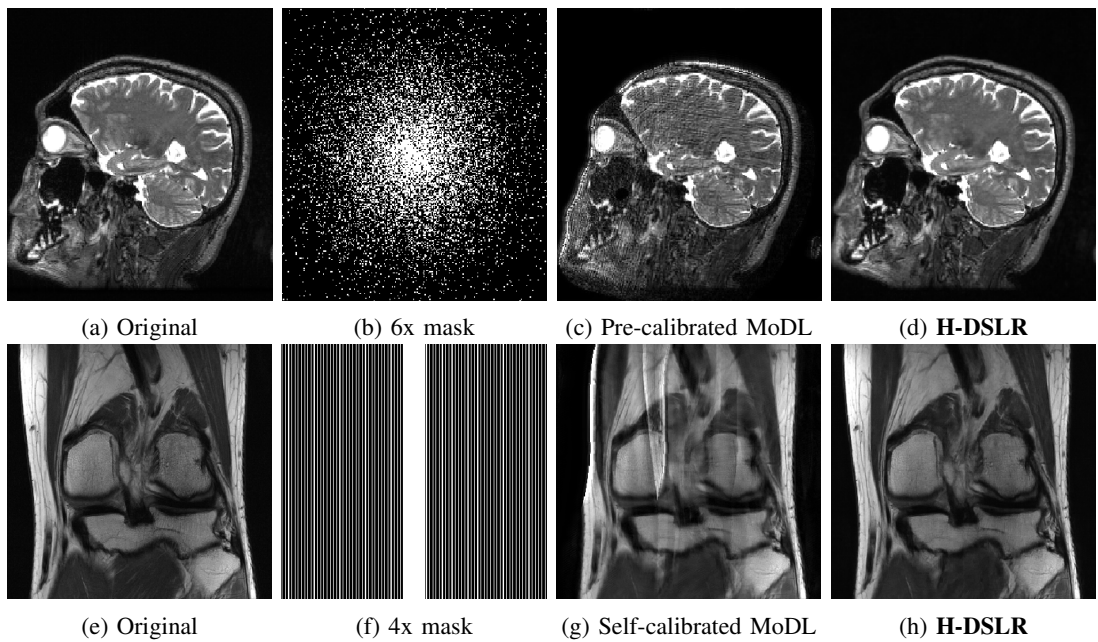}
	\caption{The top row of images (a)-(d) show comparisons of pre-calibrated MoDL with the proposed calibration-less approach during mismatches in scans. A cartesian 2D 6-fold under-sampling mask in (b) was used for under-sampling the k-space. The acquired k-space measurements were translated in spatial domain to emulate motion. The MoDL reconstruction shows diagonally striped motion artifacts due to mismatch. Our proposed scheme remains unaffected. The bottom row of images (e)-(h) display comparisons of the proposed approach with self-calibrated MoDL. The mask in (f) is used for under-sampling the k-space data and subsequent reconstruction. It samples 16 fully sampled lines in the center for calibration purposes. The coil sensitivities for MoDL are estimated using ESPIRiT \cite{uecker2014espirit} from the calibration window of 16 x 16 at the center of k-space. The performance of self-calibrated MoDL breaks down due to inaccurate sensitivities estimated from a smaller calibration region. Thus, the requirement of a larger calibration region limits acceleration. Our proposed scheme is robust to acceleration in the calibration region, thus pushing it further.}
	\label{fig:ben_trans_low_window}
\end{figure*}

Pre-calibrated approaches, which estimate coil sensitivities from calibration scans, suffer from motion-induced mismatch between the calibration and main scans, resulting in artifacts. We study the benefit of the uncalibrated deep SLR methods using a simulation. Specifically, we simulate a mismatch by modulating the k-space data of the accelerated scan with a linearly varying phase term, which corresponds to a shift in image domain. A phase shift of 5 pixels along horizontal as well as vertical direction was applied on the 2D slices, assuming a minor physical motion during scan would lead to a similar amount of shift in either direction. We compare the pre-trained MoDL and H-DSLR framework on this data, whose results are shown in Fig. \ref{fig:ben_trans_low_window}(a)-(d).  Due to the mismatch between coil images and the corresponding sensitivities, there are visible striped artifacts in the MoDL reconstruction. By contrast, we observe that the the proposed hybrid DSLR framework remains unaffected. This simulation study shows the benefit of our proposed method over calibrated setting during motion.

Self-calibrated approaches do not require an additional calibration scan and hence are not sensitive to the above motion errors. They instead leverage a fully sampled calibration region (center of k-space) to estimate the coil sensitivities. However, this approach restricts the achievable acceleration rates. While the acceleration rate can be increased by reducing the size of the calibration region, the smaller calibration region results in inaccurate sensitivity estimates. The sensitivities were estimated from a 24 x 24 region in our MoDL scheme. We now estimate the sensitivities using ESPIRiT \cite{uecker2014espirit} from a calibration window of 16 x 16. The pre-trained MoDL was tested on the dataset using those estimated sensitivities. As seen in our experimental results in Fig. \ref{fig:ben_trans_low_window}(e)-(h), the inaccurate sensitivities resulted in several visible artifacts in the MoDL reconstructions. The proposed method does not suffer from these artifacts since it is an uncalibrated scheme and does not rely on the central k-space region to estimate the sensitivities.     
 
\subsection{Comparison of the Computational Complexity}
A key benefit of the proposed Deep-SLR scheme over SLR methods is the quite significant reduction in runtime, along with the improved performance offered by the combination of the image domain prior. The recorded runtimes are shown in Table \ref{tab:run_time}. We report runtimes for 10 iterations ($K=10$) of our proposed k-space and hybrid Deep-SLR algorithms, and MoDL. We note that the DL approaches are roughly a few thousands-fold faster than the IRLS-SLR schemes in both cases. As discussed previously, SLR methods estimate the linear projection operator on the fly and require at least 50 iterations to converge. The high complexity of the SVD and the evaluation of the Gram matrix, along with the large number of iterations, is the main reason for the long runtime of the SLR methods.  By contrast, the Deep-SLR approaches pre-learn the CNNs from exemplar data, which eliminates the need for \eqref{qup}. The hybrid Deep-SLR approach is slightly slower than k-space Deep-SLR in both the cases since the former uses two CNNs compared to one by the latter even if the effective number of parameters are the same. In a single-channel setting, although K-UNET and I-UNET have more learnable parameters, these approaches are faster by virtue of a single iteration rather than multiple iterations in proposed schemes. Note that the iterative approach brings improved performance as discussed in the previous sections. In the parallel MRI setting, the Deep-SLR schemes use five-layer CNNs that make them faster than K-UNET even after multiple iterations. We note that the MoDL scheme uses a multi-channel forward model that requires a conjugate gradient (CG) algorithm to enforce DC, which makes it slower than the Deep-SLR schemes. By contrast, the proposed scheme recovers the coil images; the forward model only includes Fourier sampling, which makes these schemes faster in training and testing.
\begin{table}[t!]
	\fontsize{6}{8}
	\selectfont
	\centering
	\renewcommand{\arraystretch}{0.8}
	\begin{tabular}{ccccccc}
		\hline
		\multicolumn{6}{c}{Single-channel recovery (minutes per subject)}\\ \hline
		Organ & GIRAF & K-UNET & \textbf{K-DSLR} & I-UNET & \textbf{H-DSLR}\\ 
		Knee/Brain & 197.33 & 0.07 & 0.32 & 0.07 & 0.37\\  \hline
		\multicolumn{6}{c}{Parallel MRI recovery (minutes per subject)}\\ \hline
		Organ &PSLR & K-UNET & \textbf{K-DSLR} & MoDL & \textbf{H-DSLR} \\
		Brain &1223 & 0.7 & 0.17 & 0.83 & 0.19 \\
		Knee &3106.67 &2.83 & 0.63 &4.40 & 0.75  \\\hline
	\end{tabular}
	\vspace{1em}
	\caption{Comparison of Single-channel and Parallel MRI reconstruction times. The reported values are average reconstruction times per subject in minutes. The bold-faced methods are the proposed ones.}
	\label{tab:run_time} 
	\vspace{-2.5em}
\end{table}       
\section{Discussion and Conclusion}
We introduced a general model-based DL framework to significantly accelerate SLR matrix-completion algorithms. The key distinction with SLR methods is the pre-learning of the CNN parameters from exemplar data. Since the parameters need not be estimated from the measured data itself, the proposed algorithm is faster by several orders of magnitude. In addition, an additional image domain prior helps to further improve performance. We showed the utility of the proposed scheme in two representative applications with drastically different lifting structure. 

In most cases considered in this work, the performance of the k-space network is comparable or better than the corresponding PSLR scheme. The addition of the image domain network further improved performance. The hybrid DSLR outperforms the existing pre-calibrated MoDL scheme in the parallel MRI setting. However, the performance of the k-space DSLR scheme is marginally lower than the corresponding SLR scheme in the single-channel brain case. Additional experiments on larger datasets are needed to understand whether this is a consistent observation.  The proposed framework is applicable in theory to a wide range of SLR priors described in earlier work \cite{jacob2020structured}. In this study, we restricted our attention to three representative applications. The applicability of the proposed framework to other problem settings is beyond the scope of this work and will be considered elsewhere. The MSE was used as the loss to train the networks. Since perceptual metrics such as SSIM are related to the MSE in a non-linear fashion, the performance of the proposed networks with respect to the SSIM may be better or worse with respect to others. The training can be changed to use arbitrary loss metrics, including SSIM, which may yield more visually pleasing images than the ones trained using MSE loss. Most of the experiments in this paper were restricted to scans on the same scanners. More work is needed to determine its utility in a multi-scanner and multi-center setting. We have not addressed the design of the sampling scheme that is optimal for the problem in this work. We refer the readers to our recent work that focuses on this aspect \cite{aggarwal2020j}.

\bibliographystyle{IEEEtran}
\bibliography{refs_revision3}

\end{document}


\title{Supplementary material for \\ ``Deep Generalization of Structured Low-Rank Algorithms (Deep-SLR)"  }

\author{\IEEEauthorblockN{Aniket Pramanik,
Hemant Aggarwal,
Mathews Jacob 
}\\
\IEEEauthorblockA{The University of Iowa, USA}

\thanks{Aniket Pramanik, Hemant Aggarwal and Mathews Jacob are from the Department of Electrical and Computer Engineering at the University of Iowa, Iowa City, IA, 52242, USA (e-mail: aniket-pramanik@uiowa.edu; hemantkumar-aggarwal@uiowa.edu; mathews-jacob@uiowa.edu). This work is supported by grant NIH 1R01EB019961-01A1.}}
\IEEEtitleabstractindextext{%
\begin{abstract}
We show results from additional experiments on single-channel and parallel MRI recovery.
Later we discuss the application of our proposed schemes for diffusion MRI recovery and compare those with the state-of-the-art methods. 
\end{abstract}
}

\maketitle

\IEEEdisplaynontitleabstractindextext

\IEEEpeerreviewmaketitle
\renewcommand{\thefigure}{S\arabic{figure}}
\renewcommand{\thetable}{S\arabic{table}}
\setcounter{figure}{0}
\section{Single-channel Signal Recovery}
The proposed approaches for single-channel recovery are compared against the state-of-the-art in Table S1. The methods were tested on single-channel knee (coronal and sagittal view), and brain data mentioned in Section IV-A in the main paper. The mean SNR, PSNR and SSIM values with corresponding standard deviations were calculated for 2560 (10 subjects) brain, and 420 (3 subjects) knee (sagittal and coronal) slices, respectively. The coronal knee and brain were 4x under-sampled while the sagittal knee was 6x under-sampled. The proposed k-space network K-DSLR outperforms K-UNET (k-space UNET) and I-UNET (image-space UNET). K-DSLR performance is comparable to calibration-less approach GIRAF that motivates the proposed scheme. The addition of spatial domain prior in H-DSLR improves performance significantly over GIRAF.

\begin{figure}[h]
	\centering
	\includegraphics[width=0.5\textwidth,keepaspectratio=true,trim={1.7cm 8.2cm 11.4cm 8.8cm},clip]{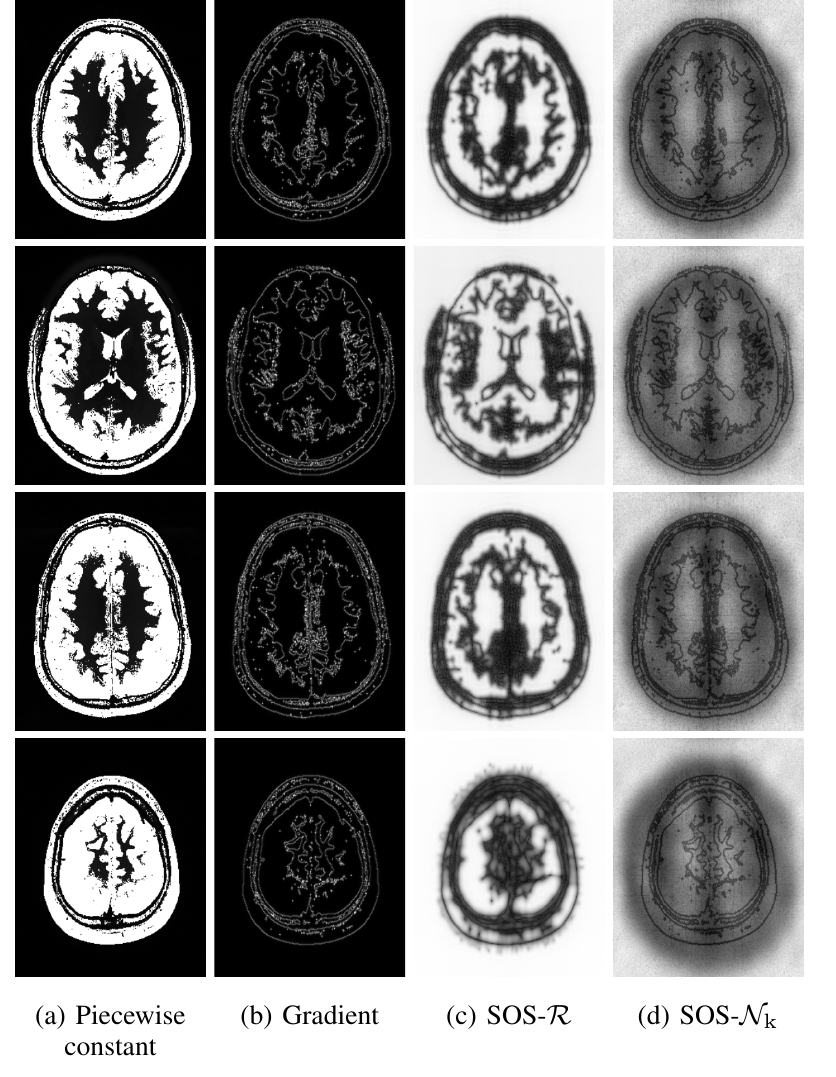}
	\caption{Illustration of linear and non-linear annihilation operators on piecewise constant images. The images and their gradients are shown in (a) and (b) respectively. The SOS-$\mathcal R$ is the sum of squares function on linear annihilation operator $\mathcal R$ from SLR schemes that kills edges or high gradient regions as shown in (c). The non-linear extension of SOS-$\mathcal R$ is SOS-$\mathcal N_{\rm k}$ and its outputs are shown in (d) which closely match to (c). The non-linear operation is generalizable across variety of brain slices shown here.}
	\label{fig:single_channel_annihilation}
\end{figure}
\begin{figure*}[h!]
	\centering
	\includegraphics[scale=0.9,keepaspectratio=true,trim={0.4cm 8.45cm 2.6cm 9cm},clip]{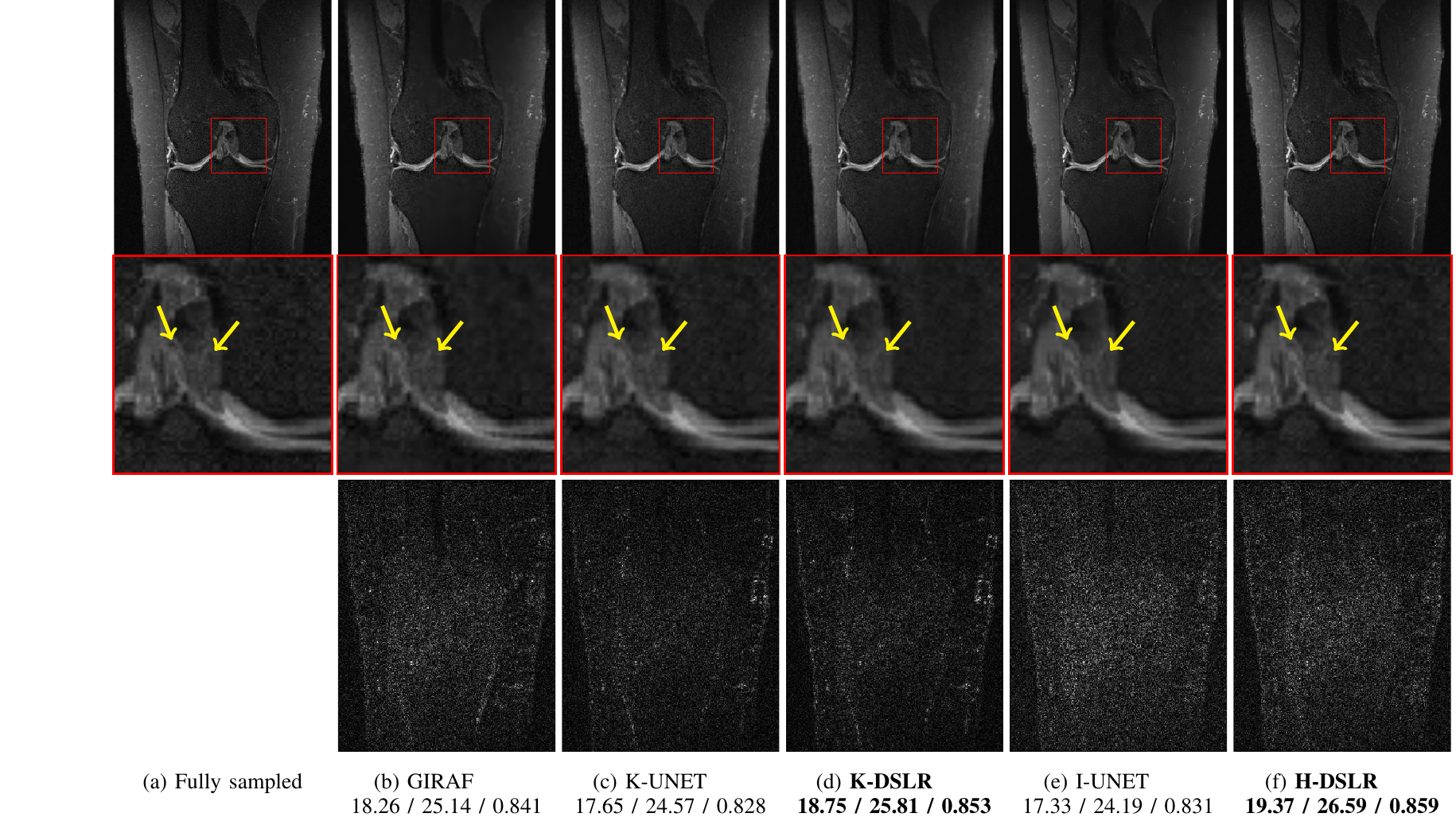}\hspace*{\fill}
	\caption{Reconstruction results of 6-fold accelerated single-channel knee data with coronal view. SNR (dB)/ PSNR (dB)/ SSIM values are reported for each case. The top row displays reconstructions (magnitude images) while the bottom row displays corresponding error maps. The yellow arrows point out the differences in the zoomed coronal view of cartilage region. The proposed schemes outperform state-of-the-art schemes and preserve complex structures better as pointed out by arrows.}
	\label{fig:knee_single_channel_zoomed_coronal}
\end{figure*}
\begin{figure*}[h!]
	\centering
	\includegraphics[scale=0.9,keepaspectratio=true,trim={1.8cm 9.1cm 2.6cm 9.7cm},clip]{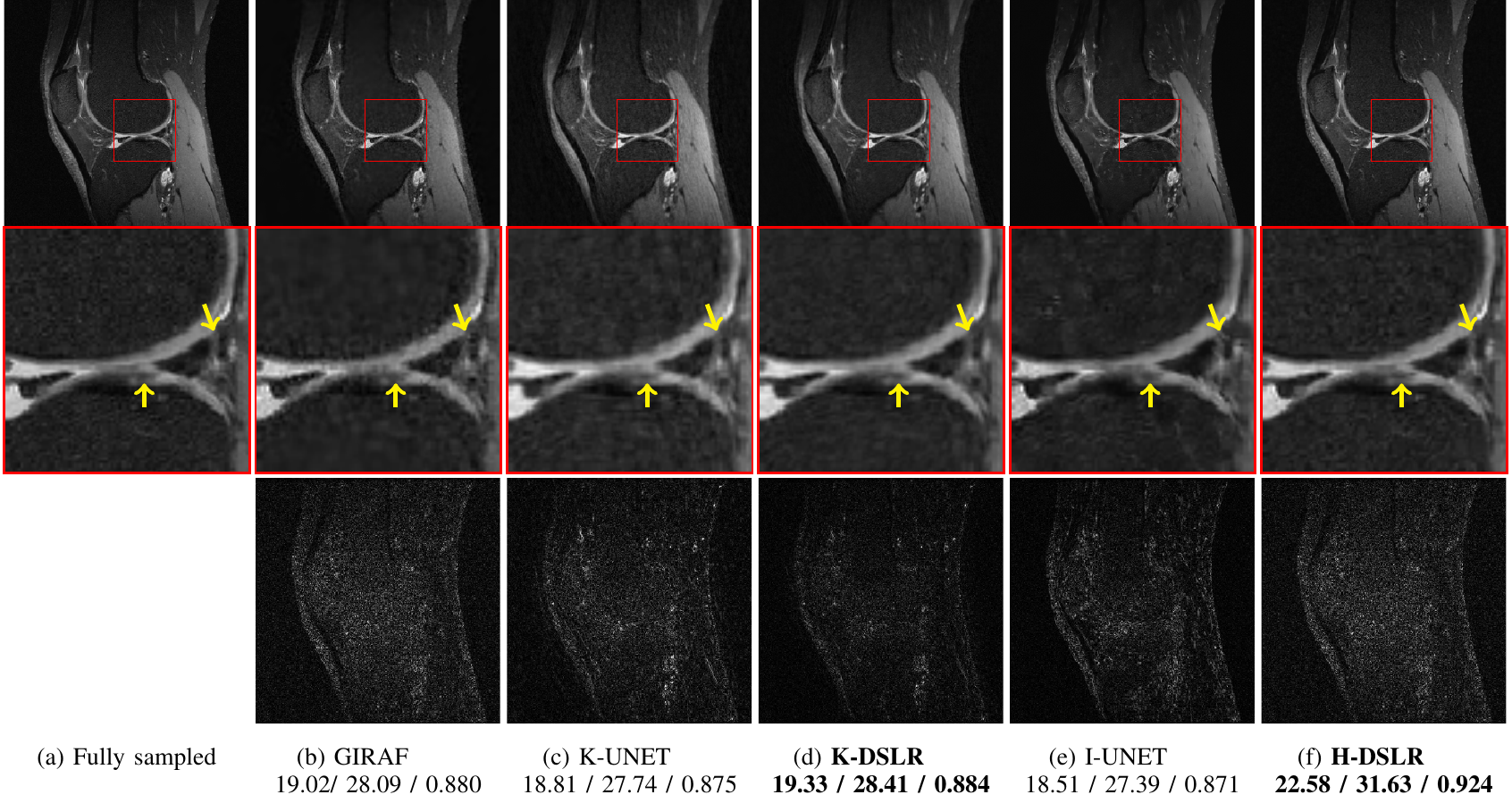}
	\caption{Reconstruction results of 4-fold accelerated single-channel knee data with sagittal view. SNR (dB)/ PSNR (dB)/ SSIM values are reported for each case. The top row displays reconstructions (magnitude images) while the bottom row displays corresponding error maps. The yellow arrows point out the differences in the zoomed sagittal view of cartilage region. The proposed schemes recover finer details better compared to others.}
	\label{fig:knee_single_channel_zoomed_sagittal}
\end{figure*}

We compare reconstruction quality of single-channel coronal and sagittal knee slices in Fig. \ref{fig:knee_single_channel_zoomed_coronal} and Fig. \ref{fig:knee_single_channel_zoomed_sagittal}, respectively. The K-DSLR and H-DSLR reconstructions are that of a $K = 10$ iteration model. We observe that K-DSLR results are comparable to model-based GIRAF. By contrast, our proposed model-based schemes outperform direct inversion approaches, K-UNET, and I-UNET. Note that our proposed schemes have much smaller number of trainable parameters compared to the UNETs. The multiple iterations of the proposed alternating strategy improves overall performance. An addition of spatial domain prior in H-DSLR visibly improves reconstruction quality, and SNR. The yellow arrows in the zoomed cartilage region point out differences in the preservation of minute structures. K-UNET and I-UNET seem to be missing lot of details as pointed out by the arrows. Although GIRAF and K-DSLR reconstructions miss few details, the H-DSLR scheme preserves all of those. The H-DSLR reconstructions are better in terms of SNR than other methods considered here. 
            
\begin{table}[h!]
	\fontsize{6}{8}
	\selectfont
	\centering
	\renewcommand{\arraystretch}{0.8}
	\begin{tabular}{|c|c|c|c|}
		\hline
		\multicolumn{4}{|c|}{Signal-to-Noise Ratio (SNR)}\\
		\hline
		Organ&\multicolumn{1}{|c|}{Knee Coronal} &\multicolumn{1}{|c|}{Knee Sagittal} & \multicolumn{1}{|c|}{Brain (CCP)} \\ \cline{2-4}
		Acceleration  & \multicolumn{1}{c}{6x} & \multicolumn{1}{|c|}{4x} & \multicolumn{1}{|c|}{4x} \\
		Methods & SNR & SNR & SNR \\ \hline 
		GIRAF &18.14 $\pm$ 1.58 &19.01 $\pm$ 1.64 & 22.02 $\pm$ 1.36 \\
		K-UNET &17.76 $\pm$ 1.29 &18.78 $\pm$ 1.54 & 19.78 $\pm$ 1.47 \\
		\textbf{K-DSLR} & 18.69 $\pm$ 1.08 & 19.16 $\pm$ 1.11 & 21.18 $\pm$ 1.01  \\
		I-UNET &17.29 $\pm$ 1.77 &18.53 $\pm$ 1.83 & 19.43 $\pm$ 1.59 \\
		\textbf{H-DSLR} & 19.31 $\pm$ 1.02 & 22.48 $\pm$ 1.26 & 25.39 $\pm$ 0.84  \\ \hline
				\multicolumn{4}{|c|}{Peak Signal-to-Noise Ratio (PSNR)}\\
		\hline
		Organ&\multicolumn{1}{|c|}{Knee Coronal} &\multicolumn{1}{|c|}{Knee Sagittal} & \multicolumn{1}{|c|}{Brain (CCP)} \\ \cline{2-4}
		Acceleration  & \multicolumn{1}{c}{6x} & \multicolumn{1}{|c|}{4x} & \multicolumn{1}{|c|}{4x} \\
		Methods & SNR & SNR & SNR \\ \hline 
		GIRAF &25.01 $\pm$ 1.55 &27.96 $\pm$ 1.59 & 29.11 $\pm$ 1.41 \\
		K-UNET &24.55 $\pm$ 1.34 &27.75 $\pm$ 1.52 & 26.68 $\pm$ 1.43 \\
		\textbf{K-DSLR} & 25.64 $\pm$ 1.02 & 28.22 $\pm$ 1.03 & 28.27 $\pm$ 0.97  \\
		I-UNET &24.34 $\pm$ 1.79 &27.49 $\pm$ 1.87 & 26.47 $\pm$ 1.63 \\
		\textbf{H-DSLR} & 26.46 $\pm$ 0.98 & 31.54 $\pm$ 1.18 & 32.53 $\pm$ 0.87  \\ \hline
		\multicolumn{4}{|c|}{Structural Similarity (SSIM)}\\
		\hline
		Organ&\multicolumn{1}{|c|}{Knee Coronal} &\multicolumn{1}{|c|}{Knee Sagittal} & \multicolumn{1}{|c|}{Brain (CCP)} \\ \cline{2-4}
		Acceleration  & \multicolumn{1}{c}{6x} & \multicolumn{1}{|c|}{4x} & \multicolumn{1}{|c|}{4x} \\
	Methods & SSIM & SSIM & SSIM \\ \hline 
		GIRAF &0.841 $\pm$ 0.031 &0.877 $\pm$ 0.040 &0.915 $\pm$ 0.023 \\
		K-UNET &0.830 $\pm$ 0.029 &0.872 $\pm$ 0.032 &0.842 $\pm$ 0.019 \\
		\textbf{K-DSLR} & 0.849 $\pm$ 0.019 & 0.878 $\pm$ 0.025 & 0.902 $\pm$ 0.017  \\
		I-UNET &0.834 $\pm$ 0.028 &0.873 $\pm$ 0.026 &0.838 $\pm$ 0.026 \\
		\textbf{H-DSLR} & 0.852 $\pm$ 0.011 & 0.921 $\pm$ 0.013 & 0.929 $\pm $ 0.009  \\ \hline	
	\end{tabular}
	\vspace{1em}
	\caption{Quantitative comparison of SLR, Deep-SLR and UNET reconstructions in the context of single-channel recovery. The bold-faced methods are the proposed ones.}
	\label{tab:comp_sc}	 
	\vspace{-2em}
\end{table}

We show results of hypothesis test (described in Section V-B in the main paper) on single-channel brain slices with different anatomies in Fig. \ref{fig:single_channel_annihilation}. The proposed k-space network learns non-linear annihilation relations that can kill edges or high gradient regions in piecewise constant images. The results in Fig. \ref{fig:single_channel_annihilation} show that the non-linear block $\mathcal N_{\rm k}$ can linearly approximate the annihilation relations which closely match with those learnt by the SLR schemes. Specifically, the SOS-$\mathcal N_{\rm k}$ outputs of the perturbations for each case in Fig. \ref{fig:single_channel_annihilation}.(d) are similar to SOS-$\mathcal R$ outputs in Fig. \ref{fig:single_channel_annihilation}.(c) from SLR schemes. Thus, the non-linear annihilation block $\mathcal N_{\rm k}$ is generalizable over a variety of brain anatomy slices. 

\section{Parallel MRI Recovery}
\begin{figure}[h!]
	\centering
	\includegraphics[scale=0.95,keepaspectratio=true,trim={1.8cm 5.4cm 11.8cm 5.8cm},clip]{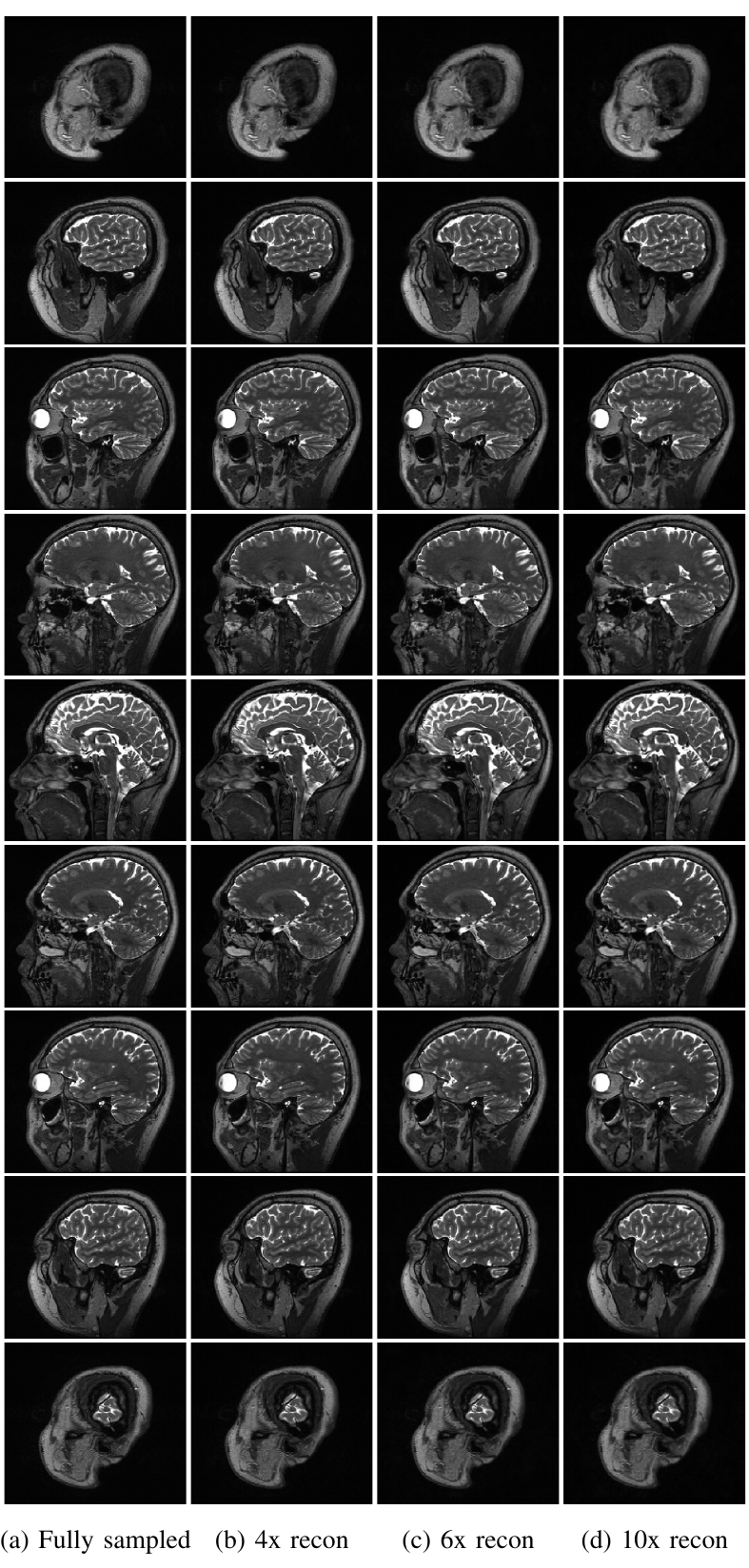}
	\caption{Proposed H-DSLR reconstructions of 12-channel brain data for 4x, 6x and 10x accelerations. The displayed images are sum-of-squares reconstructions. H-DSLR is robust to acceleration factors for different anatomies of brain.}
	\label{fig:brain_pmri_var_slices}
\end{figure}

We study robustness of the proposed H-DSLR scheme to acceleration factors for variety of slices from the test subjects. In Fig. \ref{fig:brain_pmri_var_slices}, we show reconstructions of multi-channel brain dataset for 4x, 6x and 10x accelerations over several anatomies of a test subject. We train a $K = 10$ iteration H-DSLR network end-to-end with 10x under-sampled brain slices and test it on 4x, 6x, 10x slices from subjects unseen by the network. The dataset is the one mentioned in Section IV-A from the main paper. These reconstructions are appreciable over a range of slices including the corner ones. The brain structure is preserved in all the cases; 4x and 6x reconstructions have sharper edges compared to 10x. The minute structures in the cerebellum region are better preserved in 4x compared to the other two which is due to lower acceleration. The 10x reconstruction loses few details in the cerebellum region but preserves most of the gray and white matter; 6x preserves all of them but appears slightly blurred compared to 4x. The proposed scheme could generalize appreciably over a variety of unseen brain slices at different acceleration factors.

\begin{figure}[h!]
	\centering
	\includegraphics[scale=0.9,keepaspectratio=true,trim={1.9cm 5.2cm 11.3cm 5.6cm},clip]{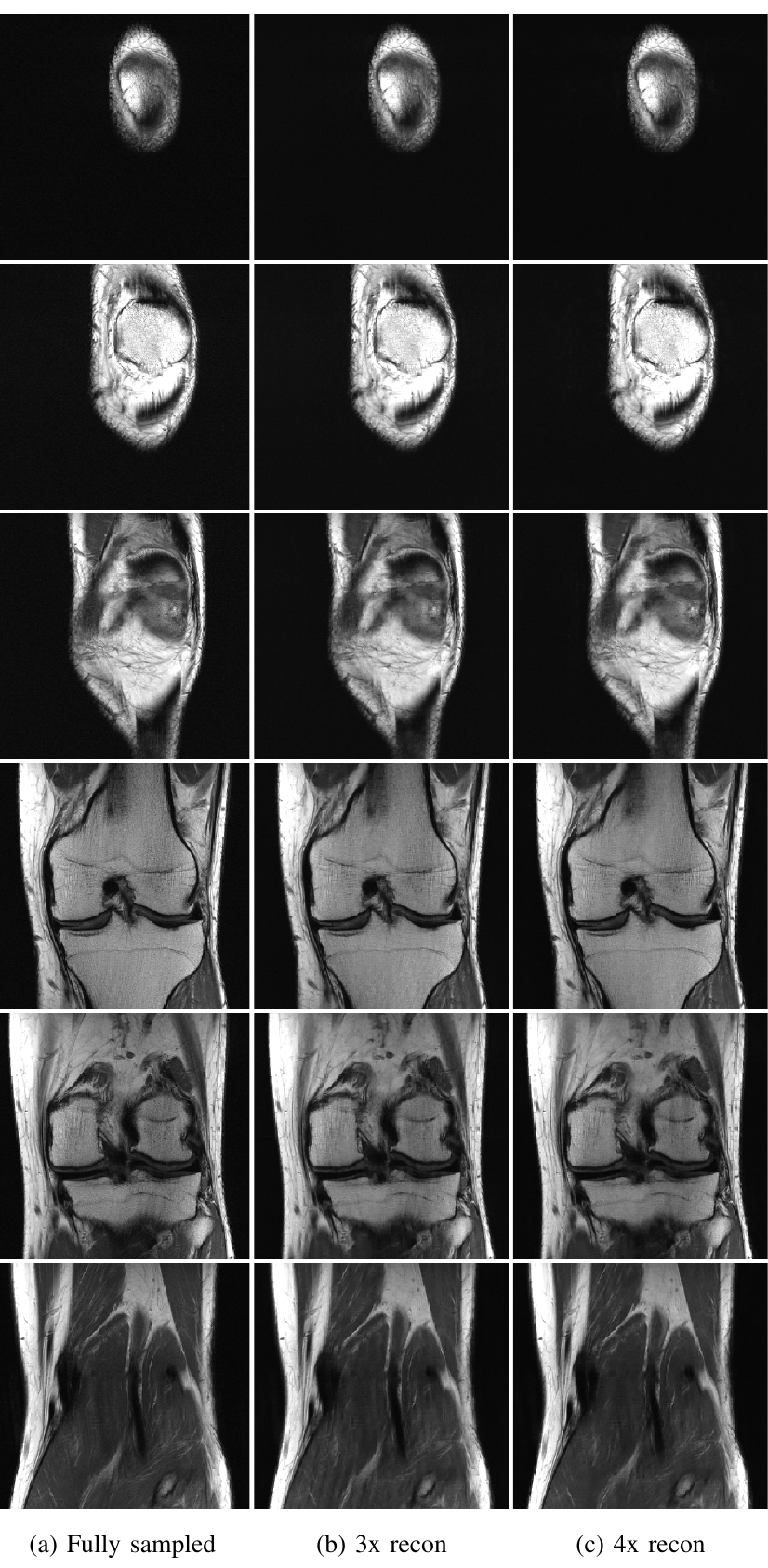}
	\caption{Proposed H-DSLR reconstructions of 15-channel knee data for 3x and 4x accelerations. The displayed images are sum-of-squares reconstructions. H-DSLR is robust to acceleration factors for different anatomies of knee.}
	\label{fig:knee_pmri_var_slices}
\end{figure}

We perform a similar study for multi-channel knee dataset described in Section IV-A in the main paper. In Fig. \ref{fig:knee_pmri_var_slices}, we display 3x and 4x reconstructions from different anatomies of a test subject. Similar to brain, we train a $K = 10$ iteration H-DSLR network with 4x under-sampled knee slices and test it on 3x, 4x slices from unseen subjects. The reconstructions appear significantly de-aliased for all cases. The cartilage region has several minute details which are slightly better preserved for the 3x case. Overall, the proposed scheme could de-alias variety of slices including the corner ones for different acceleration factors which shows its generalizability.

\begin{figure}[h!]
	\centering
	\rotatebox{90}{\hspace{20pt}Average SNR (dB) on test data}
	\includegraphics[scale=0.55,keepaspectratio=true,trim={0.5cm 0.4cm 0 0},clip]{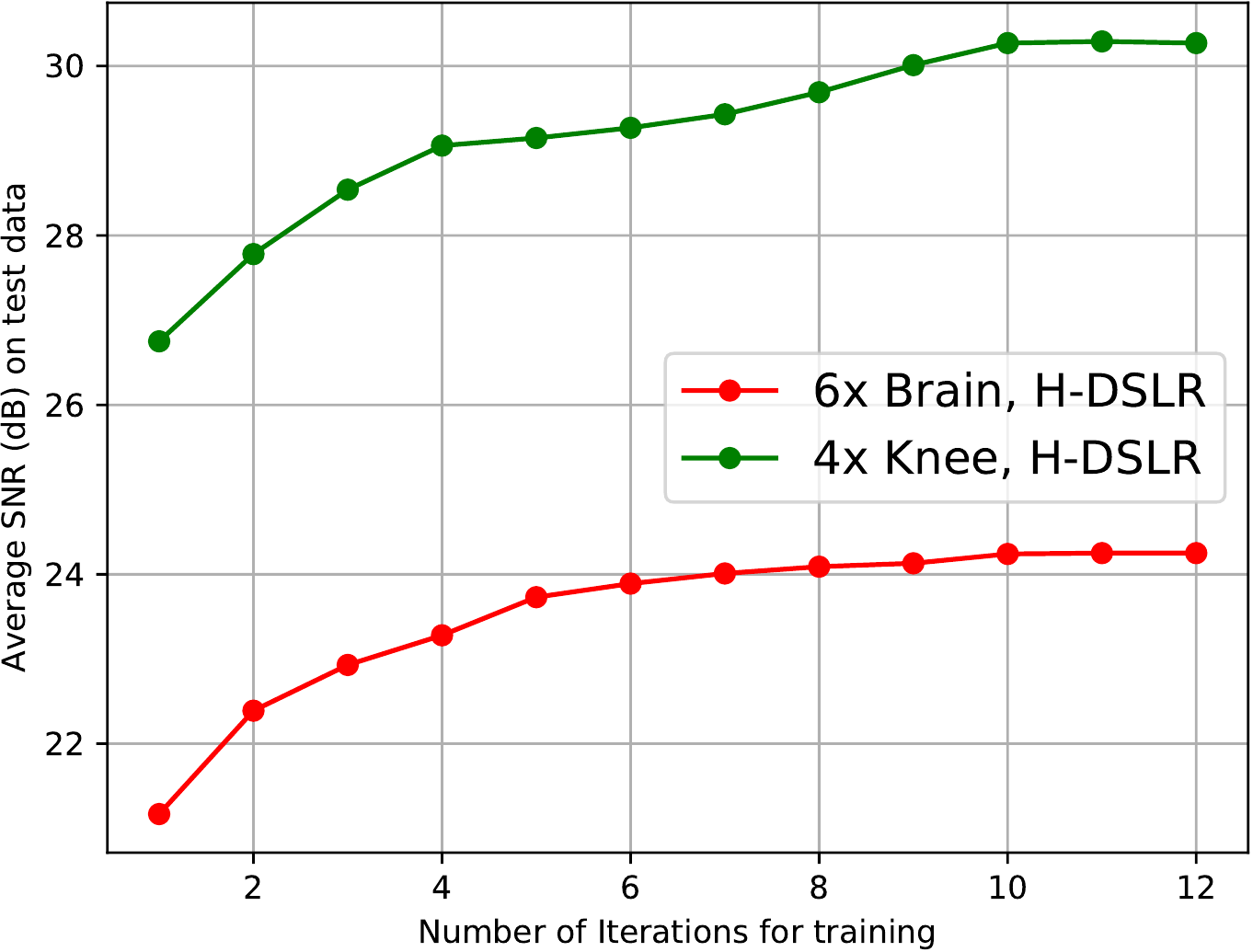}\\
	\rotatebox{0}{\hspace{20pt} Number of Iterations for training}
	\caption{Performance improvement in terms of average SNR (dB) on 6x accelerated 12-channel brain and 4x accelerated 15-channel knee test data respectively. The average testing SNR improves till $K = 10$ iteration and saturates afterwards.}
	\label{fig:ben_itr_plot}
\end{figure}

The plot in Fig. \ref{fig:ben_itr_plot} shows the effect of increasing iterations $K$ of our proposed scheme for parallel MRI cases. We observe a similar trend for 6x under-sampled brain and 4x under-sampled knee respectively. The average SNR on test data improves as we increase the iterations. Thus, unrolling the optimization blocks for several iterations is beneficial. Since, the performance saturated after $10^{th}$ iteration, we chose $K = 10$ for parallel MRI experiments. We also observed $K = 10$ iteration model to be optimal for single-channel experiments.

\begin{figure*}[h!]
	\centering
	\includegraphics[width=\textwidth,keepaspectratio=true,trim={1.7cm 11.2cm 1.6cm 11.8cm},clip]{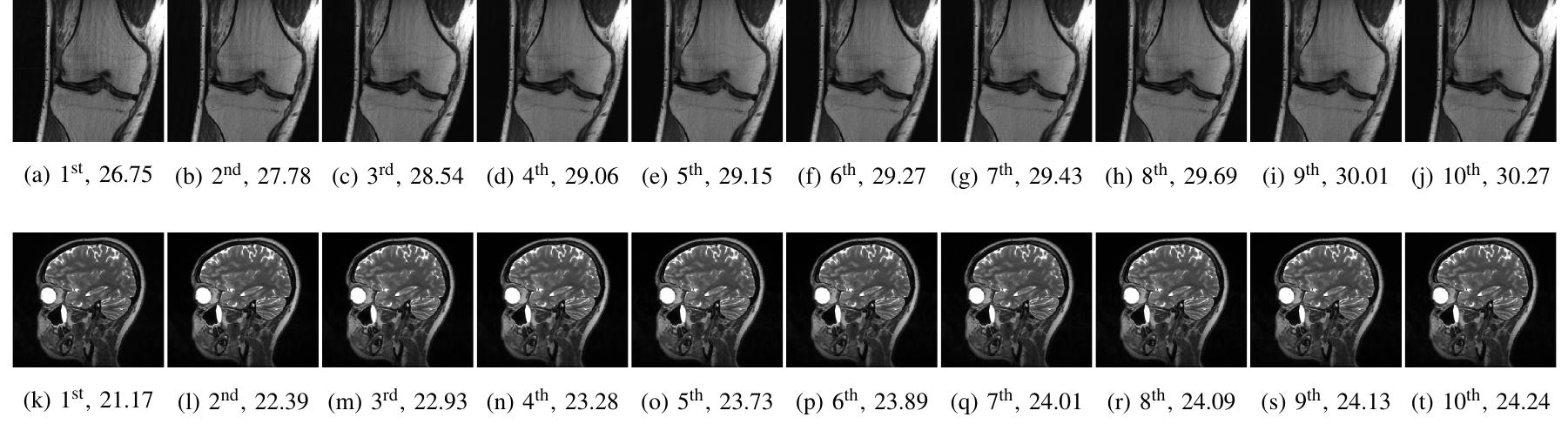}
	\caption{Proposed H-DSLR reconstructions of 4x under-sampled 15-channel knee and 6x under-sampled 12-channel brain data as a function of iterations $K$ from left to right. The images get more de-aliased with increase in iterations. The reported numbers are SNR in dB which improves with iterations.}
	\label{fig:ben_itr}
\end{figure*}

We show the intermediate results of H-DSLR algorithm as a function of iterations in Fig. \ref{fig:ben_itr}. We note that for both knee and brain test data, aliasing reduces as a function of iterations upto $K = 10$. The reduction in aliasing with more iterations justifies the benefit of unrolling the proposed scheme for more iterations. The amount of reduction in aliasing with iterations is more initially and saturates afterwards around 9-10 iterations. Thus, visible aliasing is reduced with increase in iterations which provides improved SNR.

\section{Diffusion MRI recovery}
\subsection{Structured Low Rank Algorithms for Multi-shot Echo Planar Imaging (EPI) Acquisition}
Multi-shot annihilation relations exist for phase-corrupted images as shown in \cite{mani2017multi}, in addition to the multi-channel annihilation relations discussed before. The phase-corrupted images $\boldsymbol\gamma_{i}[\mathbf r],\hspace{2pt} i=1\ldots N$ satisfy a pairwise Fourier domain annihilation relation $
\widehat{\boldsymbol \gamma_{i}}[\mathbf{k}]\ast\widehat{\boldsymbol \phi_{j}}[\mathbf{k}]-\widehat{\boldsymbol \gamma_{j}}[\mathbf{k}]\ast\widehat{\boldsymbol \phi_{i}}[\mathbf{k}]=0, \forall \mathbf{k}$, where $\widehat{\boldsymbol \gamma_{i}}[\mathbf{k}]$ and $\widehat{\boldsymbol \phi_{i}}[\mathbf{k}]$ are the Fourier coefficients of $\boldsymbol \gamma_{i}(\mathbf{k})$ and $\boldsymbol \phi_{i}(\mathbf{k})$ respectively. The $\boldsymbol \phi_{i}[\mathbf{r}]$ are smooth phase images. The relations for each pair of phase-corrupted images can be compactly written as in (5). Similar to the parallel imaging case, the Hankel matrices corresponding to each shot and channel are stacked horizontally to obtain $\mathcal T(\widehat{\boldsymbol \Gamma})$, which is low rank due to large null space $\mathbf{N}$. The columns of $\mathbf{N}$ are vertically stacked filters $\widehat{\boldsymbol \phi_{i}}$. In this case, $\mathcal G=\mathcal I$ (identity mapping). This lifting is similar to the parallel imaging case but with different annihilation relations.

\begin{figure}[h!]
	\centering
	\includegraphics[width=0.5\textwidth,keepaspectratio=true,trim={1.7cm 11.3cm 11.25cm 11.6cm},clip]{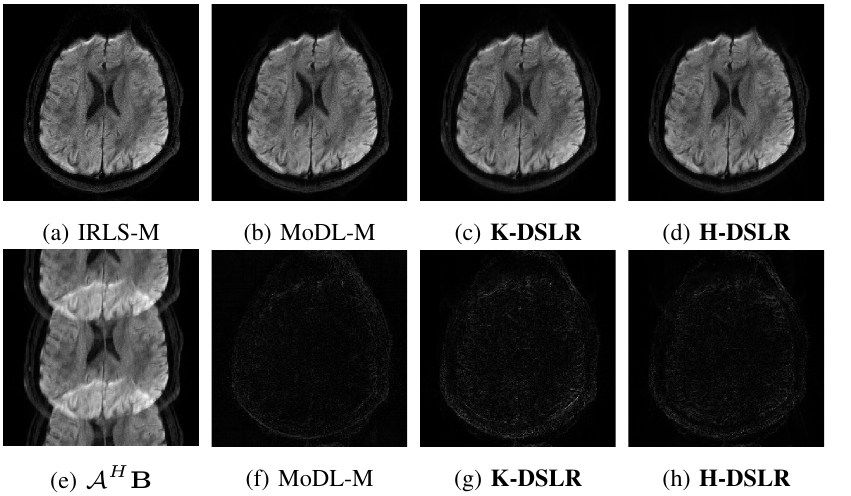}
	\caption{2-shot reconstruction results of 4-channel partial Fourier brain diffusion MRI. IRLS-M (IRLS-MUSSELS) reconstruction was ground truth for training. The proposed schemes are compared against MoDL-M(MoDL-MUSSELS). The top row shows sum-of-squares images from different schemes while the bottom row shows error maps generated from IRLS-M as ground truth. Both K-DSLR and H-DSLR provide performance comparable to MoDL-M.}
	\label{fig:diffusion_recon_2shots}
\end{figure}
\begin{figure}[h!]
	\centering
	\includegraphics[width=0.5\textwidth,keepaspectratio=true,trim={1.7cm 12.6cm 11.25cm 12.7cm},clip]{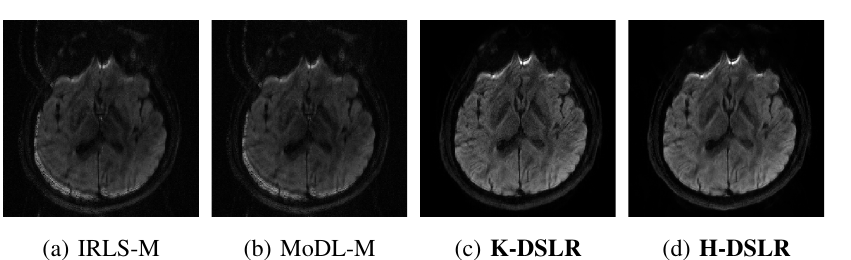}
	\caption{Comparison of calibrated approaches (MoDL-M (MoDL-MUSSELS) and IRLS-M (IRLS-MUSSELS)) with the proposed calibration-less approaches during mismatch in scans. A 2-shot recovery of the 4-channel partial Fourier brain data is shown for comparisons. The acquired k-space measurements were translated in spatial domain to emulate motion. Both calibrated approaches MoDL-M and IRLS-M reconstructions show diagonally striped motion artifacts due to mismatch. Our proposed schemes (K-DSLR and H-DSLR) remain unaffected.}
	\label{fig:ben_trans_diffusion}
\end{figure} 
\subsection{CNN Architecture for Diffusion MRI Recovery}
For this application we use the MIMO version of modified 12-layer UNET as $\mathcal N_{\rm k}$. The number of input and output channels are set according to complex channels in the dataset which is calculated as $N = N_{sh} \times N_{coil}$ where $N_{sh}$ denotes number of shots per acquisition while $N_{coil}$ corresponds to the number of coils used. Similar to other applications, we ensure the number of trainable parameters are same for both K-DSLR and H-DSLR. The regularization parameters were fixed at $\lambda = 1$, $\beta = 1$. We chose $K = 3$ iteration model based on performance which saturated with further iterations. All other training parameters were kept similar to other experiments.

\subsection{Data Acquisition}
\label{diff_data_acq}
For diffusion MRI experiments, four-shot EPI data of seven healthy subjects were obtained from \cite{aggarwal2019modl}. The subjects were scanned in a 3T scanner using 32-channel head coil. The number of gradient directions were 60 per slice with the parameters: FOV = 210 x 210 mm, TE = 84 ms, slice thickness = 4 mm and a matrix size of 256 x 152 with partial Fourier oversampling of 24 lines. The dataset was split into 68 training slices from five subjects, five validation slices from the sixth subject and six testing slices from the seventh subject. Each slice had 60 directions. Similar to \cite{aggarwal2019modl}, the IRLS-MUSSELS (IRLS-M) \cite{mani2019improved} reconstructions were used as ground truth for training and quantitative comparisons.

\subsection{State-of-the-art Methods for Comparison}
For diffusion MRI experiments, we compare our proposed scheme with MoDL-MUSSELS \cite{aggarwal2019modl} which is a non-linear extension of IRLS-MUSSELS. MoDL-MUSSELS (MoDL-M) learns non-linear Fourier domain annihilation relations along with spatial regularization. It is a phase blind recovery scheme with a pre-calibrated approach that uses coil sensitivity information estimated from additional calibration scans on top of main scan. On the other hand, our proposed method does a double (phase and coil sensitivity) blind recovery that avoids potential motion artifacts introduced from the sensitivity estimation step. Both the methods were tested on the diffusion data described in Section \ref{diff_data_acq}.

\subsection{Brain Diffusion MRI Recovery}
We performed a 2-shot recovery of brain diffusion MRI with the proposed scheme and compared against pre-calibrated MoDL-MUSSELS. All the networks were trained with IRLS-MUSSELS reconstructions as ground truth. The comparisons on one of the slices at a specific direction can be seen in Fig. \ref{fig:diffusion_recon_2shots} where error maps are generated by computing absolute differences with IRLS-MUSSELS (ground truth). The H-DSLR and MoDL-MUSSELS reconstructions look sharper at the edges compared to K-DSLR which can be attributed to the spatial prior leveraged by these schemes. K-DSLR error map shows some residual error along the skull region which are further suppressed by the spatial domain prior in H-DSLR. Proposed reconstructions are comparable to MoDL-MUSSELS visually and also through error maps. Note that MoDL-MUSSELS does a phase blind recovery by leveraging coil sensitivity information. On the other hand, the proposed schemes are calibration-less and hence perform a double blind recovery which is more challenging.

\subsection{Benefit over Calibrated Approaches}
Pre-calibrated approaches suffer from motion induced artifacts due to mismatch between the calibration and main scans. We demonstrate the benefit of our proposed calibration-less scheme over pre-calibrated MODL-MUSSELS and IRLS-MUSSELS through a simulation experiment. Similar to the parallel MRI case, we introduce a mismatch by modulating the Fourier data with a linearly varying phase term which leads to a shift in spatial domain. The reconstructions results on a test slice is shown in Fig. \ref{fig:ben_trans_diffusion}. We observe striped artifacts in both IRLS-MUSSELS and MoDL-MUSSELS reconstructions due to a mismatch between the sensitivities and coil images while our proposed calibration-less approaches remain unaffected. This study shows the benefit of our proposed scheme in avoiding motion artifacts over calibrated approaches.

\bibliographystyle{IEEEtran}
\bibliography{refs_revision3}